\documentclass[11pt]{article}

\usepackage[preprint]{acl}

\usepackage{times}
\usepackage{latexsym}

\usepackage[T1]{fontenc}

\usepackage[utf8]{inputenc}

\usepackage{microtype}

\usepackage{inconsolata}

\usepackage{graphicx}
\usepackage{booktabs}
\usepackage{multirow}
\usepackage{fontawesome5}
\usepackage[most]{tcolorbox}
\usepackage{xcolor}
\usepackage{enumitem}

\newlist{boxdialogue}{description}{1}
\setlist[boxdialogue]{
  labelwidth=4.5em,
  leftmargin=5em,
  itemsep=3pt,
  parsep=0pt,
  topsep=2pt,
  partopsep=0pt,
  font=\normalfont
}
\usepackage{soul}
\usepackage{xcolor}
\sethlcolor{yellow}
\usepackage{amsmath}
\usepackage[table]{xcolor}
\usepackage{adjustbox}
\usepackage{arydshln}

\title{ODRA: Synthesizing Cognitive Behavioral Therapy Sessions \\with Structured Chain-Of-Thought and Dynamic Patient Resistance}

\author{
 \textbf{Javier Rodriguez-Juan\textsuperscript{1}},
 \textbf{Hiba Arnaout\textsuperscript{2}},
 \textbf{Jose Garcia-Rodriguez\textsuperscript{1}},
\\
 \textbf{David Tomás\textsuperscript{1}},
 \textbf{Iryna Gurevych\textsuperscript{2}}
\\
\\
 \textsuperscript{1}3DPLab, Department of Computer Science and Technology, University of Alicante
\\
 \textsuperscript{2}UKP Lab, Department of Computer Science and Hessian Center for AI (hessian.AI), 
\\
  Technische Universität Darmstadt
\\
 \small{
   \textbf{Correspondence:} \href{mailto:j.rodriguezjuan@ua.es}{j.rodriguezjuan@ua.es}
 }
}

\begin{document}

\maketitle

\begin{abstract}
Synthetic generation of Cognitive Behavioral Therapy (CBT) sessions is challenged by two competing demands: adhering to strict therapeutic structure while modeling the resistant, unpredictable behavior of real patients. Existing script-based methods fail to capture dynamic therapeutic interactions, while multi-agent approaches struggle to adhere to CBT's sequential structure; both suffer from sycophancy, producing overly compliant patients that misrepresent real clinical settings. In this work we introduce ODRA, a novel framework for synthesizing therapy dialogues through a Chain-of-Thought (CoT) strategy grounded in CBT guidelines \cite{beck2020cognitive}. ODRA further incorporates a resistance orchestrator to solve patient sycophancy, which employs steering techniques to elicit behaviors aligned with their resistance level. Automated and expert evaluations show that ODRA significantly outperforms existing methods across therapeutic skills, CBT alignment, and patient behavioral fidelity, with licensed psychologists preferring ODRA sessions across 12 of 13 clinical metrics. Furthermore, models fine-tuned on our dataset demonstrate superior therapeutic performance against both cooperative and resistant patients, validating that explicit resistance modeling in synthetic training data directly translates to downstream clinical robustness.\footnote{Code is available at \url{anonymous.4open.science/r/ODRA}, and data is provided as supplementary material.}


\end{abstract}
\section{Introduction}
The prevalence of mental health disorders has increased significantly in recent years \cite{foulkes-2023-mhealth,McGorry2024}. While Large Language Models (LLMs) offer a promising avenue for scaling mental health support \cite{Stade2024,nguyen-etal-2025-large,na-etal-2025-survey}, their development is severely constrained by a scarcity of high-quality data, largely due to strict privacy and ethical requirements \cite{badawi-etal-2026-trust}. Although synthetic data generation has emerged as a promising solution \cite{bao-etal-2023-synthetic,Giuffre2023}, existing methods often struggle to produce faithful data aligned with real-world therapeutic discourse, where therapists follow clinical protocols, and patients exhibit a wide spectrum of different behaviors \cite{bn-etal-2025-real}.

\begin{figure}[t]
  \includegraphics[width=\columnwidth]{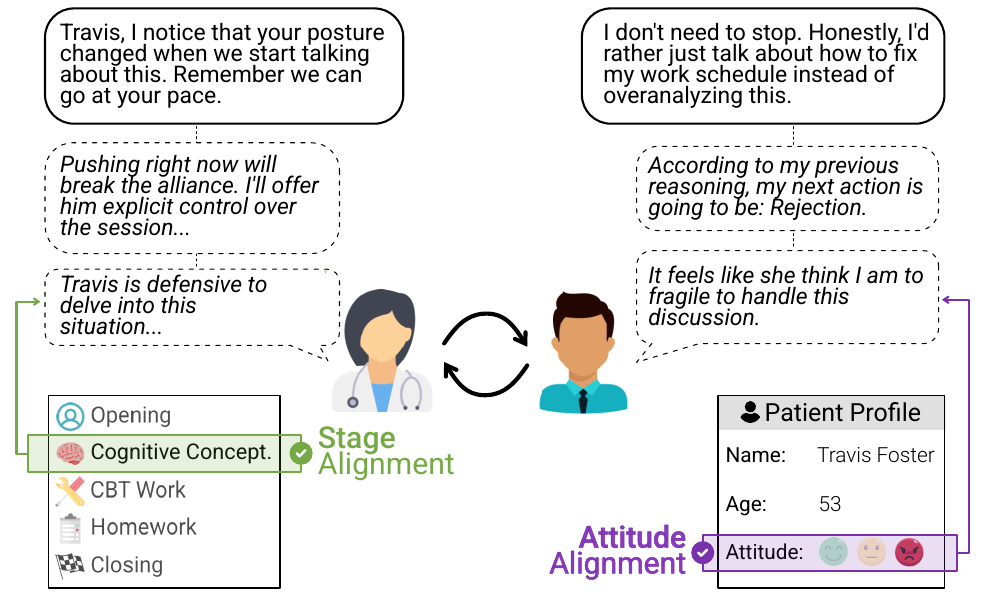}
  \caption{The ODRA framework synthesizes CBT counseling sessions using a Chain-of-Thought approach that ensures strict adherence to the sequential stages defined in foundational CBT guidelines, while eliciting resistant patient behaviors aligned with their assigned profiles.}
  \label{fig:ga}
\end{figure}

Recent work aims to mitigate this misalignment by employing script-based \cite{lee-etal-2024-cactus, kim-etal-2025-mirror}, or multi-agent methods \cite{mandal2026magnet,yang-etal-2025-cami}. Script-based approaches synthesize entire counseling sessions from a single generation pass, excelling at following structural guidelines while failing to model dynamic therapeutic interactions. Conversely, multi-agent methods foster realistic conversational flow by alternately generating therapist and patient utterances, but struggle to precisely adhere to structured clinical protocols. Additionally, they neglect to model challenging patient behaviors—often referred to as \emph{resistance}—that obstructs the therapeutic process, such as defensiveness or dismissive speech \cite{westra2018patientresistance,Chapman2016tibs}. These problems are exacerbated when simulating highly structured frameworks like Cognitive Behavioral Therapy (CBT) \cite{beck2020cognitive}. CBT relies on a progression of sequential stages to alleviate psychological distress by modifying maladaptive cognitive patterns. Due to its broad applicability \cite{cuijpers2019psychotherapies}, this protocol is frequently used as a reference to generate synthetic therapy dialogues \cite{lee-etal-2024-cactus, mandal2026magnet}.

Despite its popularity, we identify three main gaps in existing CBT synthetic methods: \textbf{(1) Structural inaccuracy.} While current methods incorporate CBT techniques, they lack a holistic framework covering all the stages defined in established CBT protocols \cite{beck2020cognitive}. \textbf{(2) Patient Sycophancy.} Existing methods frequently suffer from over-agreeability \cite{sharma2025sycophancy}. Without robust mechanisms to model patient resistance, these models elicit unrealistic behaviors that fail to simulate the challenges therapists face in real-world practice. \textbf{(3) Unvalidated Reasoning Trajectories.} Although Chain-of-Thought (CoT) approaches have been recently explored \cite{chen-etal-2025-catch}, there is no existing method providing clinically-validated, CBT-grounded reasoning traces suitable for internalizing expert decision-making into therapeutic models (see Figure \ref{fig:ga}). We address these gaps with the following contributions:
\vspace{-0.15cm}
\begin{enumerate}
    \vspace{-0.4cm}
    \item We introduce ODRA, the first generation framework \textbf{strictly aligned with the foundational CBT protocol}. By employing a CoT objective-driven approach, our method replicates the structural dynamics of full therapeutic sessions while providing clinically-validated reasoning trajectories that capture the underlying logic of expert interventions.
    \vspace{-0.2cm}
    \item We propose a novel \textbf{resistance profiling mechanism using steering techniques} to elicit challenging patient behaviors, accurately reflecting difficulties posed by resistant patients in authentic clinical settings.
    \vspace{-0.2cm}
    \item We conduct \textbf{extensive evaluations} of our framework utilizing \textbf{both automated metrics and human expert assessment}. Results show that ODRA improves text-only methods in therapist counseling skills by +10.25\%, enhances patient behavioral fidelity by +63.87\%, and is ranked by experts as the preferred generation method across multiple clinical adherence and patient realism metrics.
    \vspace{-0.2cm}
    \item We \textbf{fine-tune Llama-3 and Qwen-3.5 models} to compare the \textbf{downstream utility} of ODRA dialogues, showing that ODRA fine-tuned models outperform text-only baselines in counseling skills, achieving gains of +26.20\% and +17.02\% when interacting with cooperative and resistant patients, respectively.
    \vspace{-0.2cm}
    \item We \textbf{release our fine-tuned models and a dataset} composed of 150 CBT sessions, consisting of 9,577 turns and a total of 18,496 reasoning traces.
\end{enumerate}
\section{Related Works}\label{sec:sota}

\textbf{Cognitive Behavioral Therapy}. CBT focuses on identifying and restructuring cognitive distortions, which are defined as negative biases in thinking driven by individual's core beliefs \cite{kuru2018distortions,sharma2023cognitive,hotta-etal-2025-metamo}. These distortions are often derived from cognitive models, which are structures that describe how people's thoughts influence their emotional and behavioral reactions \cite{chahar2020associations,wang-etal-2024-patient}. The foundational CBT protocol relies on strict structural elements—such as mood checks or homework reviews—which are organized into sequential stages to guide the therapeutic process.

\noindent
\textbf{Synthetic counseling session generation.} Generation methods can be broadly categorized into script-based \cite{mandal2026graph2counsel,lee-etal-2024-cactus,zhang-etal-2024-cpsycoun} and multi-agent architectures \cite{mandal2026magnet,yang-etal-2025-cami,nguyen2026calmit,xiao-etal-2024-healme,vu2025roleplaying}. Script-based approaches often rely on pre-session outlines derived from patient input data to synthesize conversations using a single prompt \cite{lee-etal-2024-cactus}. Existing work explores diverse patient profiling inputs, ranging from intake forms \cite{lee-etal-2024-cactus} and questionnaires \cite{vu2025roleplaying} to psychological graphs \cite{mandal2026graph2counsel}. For instance, Graph2Counsel \cite{mandal2026graph2counsel} leverages these graphs to foster realism by incorporating deeper clinical traits, such as interactions between thoughts and emotions. However, the reliance of this method on psychological graphs constrains its scalability. Furthermore, while script-based methods are effective for adherence to clinical protocols, their rigid nature makes them less capable of simulating dynamic therapeutic interactions. MAGneT \cite{mandal2026magnet} solves this problem by proposing a multi-agent framework where the therapist and the patient alternately generate utterances. This setup promotes realistic conversational flow but degrades CBT structural alignment, as LLMs frequently struggle to adhere to clinical guidelines over long multi-turn interactions \cite{tan2026multiagentadherence}. Recent works have incorporated CoT steps to model latent clinical reasoning \cite{xu2025reasoningsurvey,kim-etal-2025-multimodal,hu-etal-2025-psyadvisor, chen-etal-2025-catch}. However, none of these methods address the generation of full CBT encounters. Our work addresses these limitations by providing a multi-agent architecture that faithfully captures the multi-stage progression of complete CBT sessions.


\noindent
\textbf{LLM Sycophancy.} A pervasive issue in LLM-based multi-agent interactions is sycophancy \cite{sharma2025sycophancy,pitre-etal-2025-consensagent,hong-etal-2025-syconbench}, a behavior in which agents reinforce each other's outputs rather than maintaining their assigned personas \cite{bn-etal-2025-real}. In a therapeutic context, this manifests as over-agreeability, which hinders dialogue fidelity when the patient is supposed to express negative attitudes. In MIRROR \cite{kim-etal-2025-mirror}, a static resistance category is included in the patient profile to model patient stance. Nevertheless, relying on static variables limit the ability to capture resistance dynamics throughout the session. CALM-IT \cite{nguyen2026calmit} address this issue by incorporating a dynamic resistance variable, however, this is utilized within a simple prompting strategy that lacks attitudinal instructions, making it insufficient to shift patient attitudes. To overcome these limitations, we propose the usage of LLM steering techniques, whose aim is to align model's outputs with input concepts via parametric methods \cite{wu2026improved,cao2024personalized,silva-etal-2025-steering}, or prompting strategies, which efficacy was demonstrated in recent works \cite{wu2025axbench,banayeeanzade2025psychological}. In contrast to the CALM-IT approach, prompt steering techniques include specific guidelines on how the model should behave. Despite their effectiveness, steering has not yet been explored for modeling patients within counseling sessions. To bridge these gaps, ODRA incorporates a mechanism to produce resistance updates, while employing prompt steering techniques to reflect that resistance in patient utterances.

\section{ODRA}\label{sec:odra}

\begin{figure*}[t]
  \centering
  \includegraphics[width=1\linewidth]{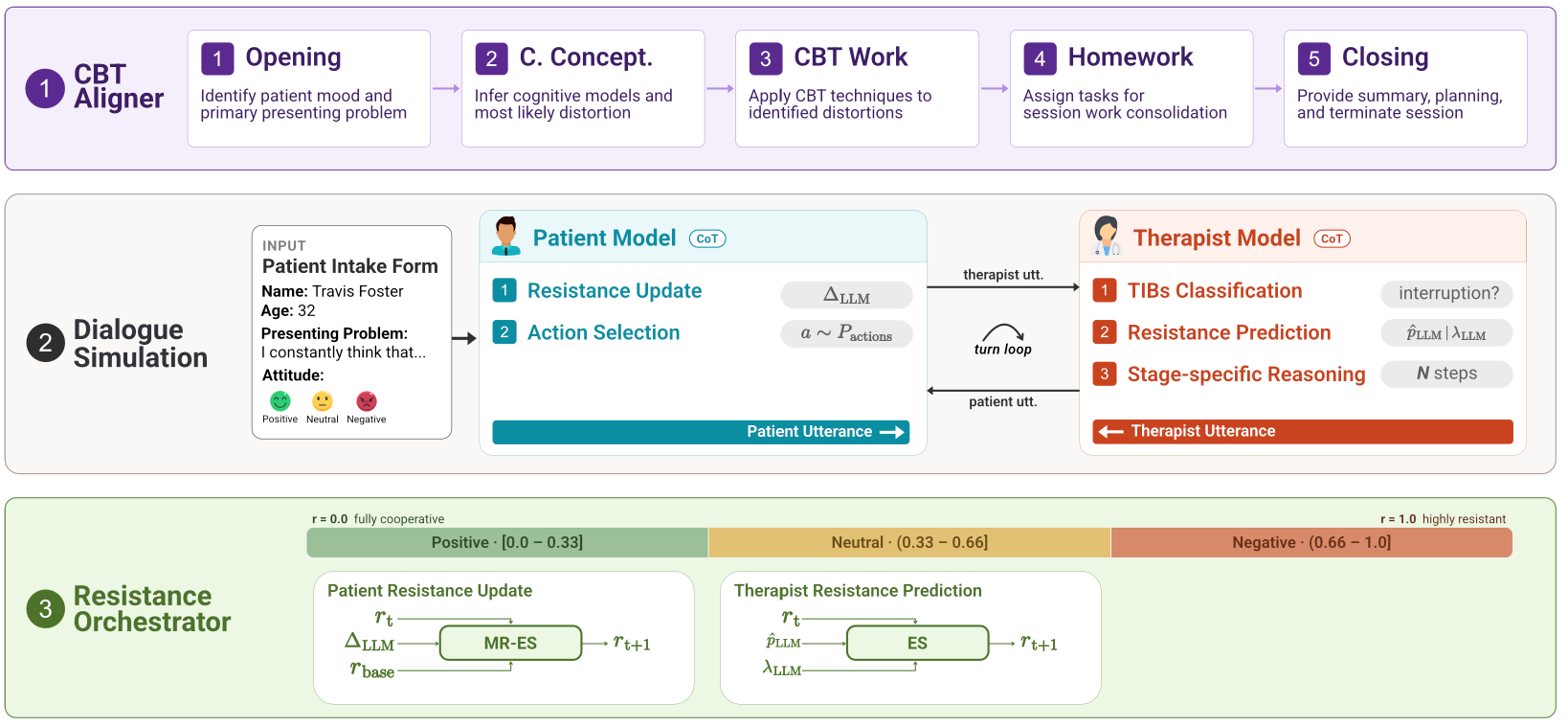}
  \caption{Architectural overview of ODRA. (\textbf{A}) The CBT Aligner controls session execution across a five-stage pipeline. (\textbf{B}) The intake form models patient profile and attitude. The patient updates resistance in response to therapist input, samples an action, and generates an utterance guided by the action, updated resistance, and steering prompt. On the other side, the therapist evaluates patient utterances for Therapy-Interfering Behaviors (TIBs), predicts patient resistance, and generates responses through stage-specific reasoning. (\textbf{C}) The Resistance Orchestrator employs Exponential Smoothing (ES) or Mean-Reverting ES (MR-ES) to update resistance, while the Behavioral Profiler (BP) maps resistance to an attitudinal steering prompt and action distribution.}
  \label{fig:odra}
\end{figure*}

Figure \ref{fig:odra} shows the main components of ODRA: The CBT Aligner, the Resistance Orchestrator, and both the therapist and patient models.




\subsection{CBT Aligner}\label{ssec:cbt}
To model the structural dynamics of real-world CBT, we developed a multi-stage framework grounded in foundational CBT guidelines \cite{beck2020cognitive}. Since a standard CBT therapeutic process consists of multiple sessions, we adapted this methodology for a single-session synthesis paradigm. This adaptation results in a five-stage pipeline: \textbf{(1) Opening.} An initial stage aimed at identifying patient mood and primary presenting problem; \textbf{(2) Cognitive Conceptualization.} A discovery stage where the therapist identifies cognitive models (see Section \ref{sec:sota}). For each cognitive model, the therapist infers the most likely associated cognitive distortion (see Appendix \ref{apx:beck-theory}); \textbf{(3) CBT Work.} The core stage where CBT techniques are employed to address identified cognitive distortions; \textbf{(4) Homework.} A consolidation stage where the patient receives homework assignments; \textbf{(5) Closing.} The final stage where the therapist provides session insights, planning, and terminates the session. Appendix \ref{apx:beck-adaptations} shows single-session synthesis adaptations, and Appendix \ref{apx:cbtgrounder-details} presents further implementation details from this component.

Throughout these stages, the framework generates stage-specific intermediate outputs—such as therapeutic plans and stage completion checks—prior to generating the final therapist utterance (see Figure \ref{fig:cbt-aligner}). We capture these intermediate outputs as CoT reasoning traces, since they encapsulate the latent CBT logic guiding the therapist decision-making. Furthermore, our framework incorporates state management to save important information across the simulation. For instance, it collects the cognitive models extracted during \emph{Cognitive Conceptualization}, and utilizes them in \emph{CBT Work} to guide the selection of techniques.

\subsection{Resistance Orchestrator}\label{ssec:resistance}
To increase the fidelity of patient behaviors, we designed a mechanism that simulates resistance to therapy. We model patient resistance as a continuous dynamic variable $r \in [0.0, 1.0]$, bounding the behavior between completely cooperative ($r=0.0$) and highly resistant ($r=1.0$). This module performs turn-by-turn updates for patient resistance and therapist estimated resistance. While the patient resistance represents the ground-truth value, the therapist estimated counterpart is a prediction of the most likely patient resistance. This is used to adapt therapist utterances to the patient internal state. This component comprises three main modules: The Patient Resistance Update, the Therapist Resistance Update and the Behavioral Profiler.


\textbf{Patient Resistance Update.} On the patient side, resistance updates are conditioned on the patient previous resistance state, their baseline resistance, and their perceived helpfulness of the therapist last utterance. To quantify this perception, we prompt the patient model to infer a resistance shift ($\Delta_{\text{LLM}}$), which represents the change in their resistance from the previous turn (see Appendix \ref{apx:prompts} for the prompt). To prevent unrealistic resistance fluctuations, the model bounds this shift within $[-0.10, 0.15]$, with boundaries determined via hyperparameter tuning (see Appendix \ref{apx:rdelta-tuning}). We employ asymmetric bounds to align with the negativity bias in human cognition, where negative emotions produce a stronger psychological impact than positive ones \cite{baumeister2001bad}. This means that patients are more prone to experience greater increases in resistance than decreases within a single turn. Additionally, changes in human emotions are influenced by emotional homeostasis \cite{von2024homeostasis}, a psychological principle where internal mechanisms auto-regulate affective states to restore baseline equilibrium. To computationally model this phenomenon, we implement a mean-reverting exponential smoothing function \cite{barrow2020smoothing}, which is defined as:

\vspace{-4mm}
\begin{equation}
    r_{\text{t+1}} = r_{\text{t}} + \Delta_{\text{LLM}} - \alpha (r_{\text{t}} - r_{\text{base}})
\end{equation}

\noindent where $r_{\text{t}}$ is the resistance at turn $t$, $\Delta_{\text{LLM}}$ is the inferred resistance shift, $r_{\text{base}}$ is the baseline resistance, and $\alpha$ is the homeostatic reversion rate. We set $\alpha = 0.15$ based on empirical tuning, as this provides an optimal balance between homeostatic stabilization and responsiveness to therapeutic dynamics.
(see Appendix \ref{apx:rdelta-tuning}).

\textbf{Therapist Resistance Update.} On the therapist side, we adopt a similar approach in which the therapist model infers two distinct parameters: an estimated resistance target ($\hat{p}_{\text{LLM}}$) and a clarity signal ($\lambda_{\text{LLM}}$), both bounded within $[0.0, 1.0]$. The resistance target represents the therapist estimation of the patient absolute resistance, while the clarity signal serves as a confidence metric. During an initial six-turn calibration window—which matches the maximum duration of the Opening stage (Section \ref{ssec:cbt})—the therapist employs a flexible update function to rapidly adjust the estimated resistance to the new absolute observation. This enables the model to calibrate effectively toward the ground-truth patient resistance. The therapist estimated resistance during this window is based on an exponential smoothing defined as:

\vspace{-4mm}
\begin{equation}
    p_{\text{t+1}} = p_{\text{t}} + \lambda_{\text{LLM}} (\hat{p}_{\text{LLM}} - p_{\text{t}})
    \label{eq:therapist-initial}
\end{equation}

\noindent where $p_{\text{t}}$ is the estimated resistance at turn $t$. For the remainder of the session, this update function incorporates a post-hoc clipping mechanism. This mechanism clamps the state update within the bounds $[-0.10, +0.15]$, ensuring behavioral coherence with patient updates and preventing hallucinated prediction spikes. The update function for the rest of the session is defined as:

\vspace{-5mm}
\begin{equation}
    p_{t+1} = p_t + \left[ \lambda_{\text{LLM}} (\hat{p}_{\text{LLM}} - p_{\text{t}}) \right]_{\delta_{\min}}^{\delta_{\max}}
\end{equation}

\noindent where the hyperparameter bounds are symmetric with patient updates ($\delta_{\min} = -0.10, \delta_{\max} = 0.15$). Prompts used to infer resistance parameters are provided in Appendix \ref{apx:prompts}.

\textbf{Behavioral Profiler.} Building upon CACTUS \cite{lee-etal-2024-cactus}, we established three categorical attitudes which we map to a specific interval across the $[0.0, 1.0]$ resistance range: $\text{Positive} \in [0.0, 0.33]$, $\text{Neutral} \in (0.33, 0.66]$, and $\text{Negative} \in (0.66, 1.0]$. For profile initialization, we use the exact midpoint of its corresponding interval (i.e., $0.165$, $0.495$, and $0.83$, respectively). Unlike baseline methods that rely on static prompting, we implement a dynamic prompt steering that systematically aligns the patient utterances with the targeted attitude (see Appendix \ref{apx:prompts} for prompts). Prior to generating an utterance, this mechanism evaluates the patient updated resistance and selects the corresponding attitudinal prompt matching the active interval. Each steering prompt includes four main components: \textit{General Rules:} Coarse-grained instructions that broadly steer model's behavior toward the targeted attitude. \textit{Calibration Rules:} Fine-grained instructions that adjust model's tone to its precise resistance scalar. \textit{Should Not Rules:} Guardrails that constrain the model from exhibiting attitudes outside its active interval. \textit{In-Context Learning (ICL):} Demonstrations providing patient utterances aligned with the expected attitude. To construct our attitudinal prompts, we adopted the pipeline from \citet{wu2025axbench}, which leverages an LLM to synthesize instructions from user guidelines (see Appendix \ref{apx:prompts} for details).


\subsection{Therapist \& Patient models}\label{ssec:therapist}
\begin{figure}[t]
  \includegraphics[width=\columnwidth]{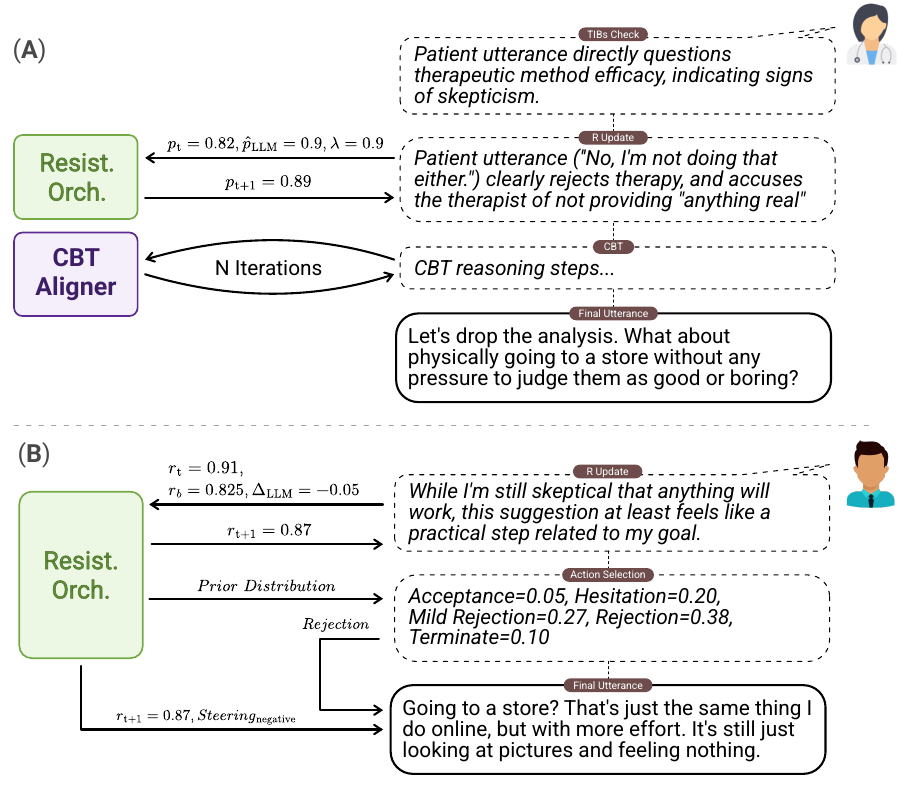}
  \caption{Execution example of Therapist and Patient CoT and utterance generation. (\textbf{A}) The therapist checks for TIBs, updates its estimated resistance, executes $N$ stage-specific reasoning steps (see Figure \ref{fig:cbt-aligner}), and generates its utterance. (\textbf{B}) The patient updates its internal resistance, uses prior distribution to sample an action, and employs this sampled action, the new resistance and the steering prompt to produce its new utterance.}
  \label{fig:models}
\end{figure}

Both the therapist and the patient are conditioned on a CoT process composed of different steps (see Figure \ref{fig:models}). On the therapist side, the CoT involves three steps: Therapy-Interfering Behavior (TIB) classification \cite{Chapman2016tibs}, estimated resistance prediction (Section \ref{ssec:resistance}), and stage-specific therapeutic reasoning (Section \ref{ssec:cbt}). The first step identifies TIBs, which are defined as patient behaviors that significantly interfere with therapy. When a TIB is detected, the therapist temporarily interrupts the CBT process to address it in the next utterance (see Figure \ref{fig:tib-detection-prompt}).


On the patient side, the CoT process consists of two steps: resistance update (Section \ref{ssec:resistance}), and the action selection. To diversify patient responses, actions are sampled from a distribution obtained by fusing a prior distribution dependent on the patient resistance interval with a contextual distribution inferred within the patient CoT process (see Figure \ref{fig:action-probability-prompt}). Subsequently, the patient employs the sampled action, the attitudinal prompt from the behavioral profiler, and the textual reasoning from the latest resistance update to generate their next utterance.

\section{Experimental setup}\label{sec:exp-setup}

\begin{table*}[htpb]
  \centering
  \small 
  \setlength{\tabcolsep}{5pt} 
  \renewcommand{\arraystretch}{1.5}
  
  \newcommand{\bluebg}{\cellcolor{blue!10}}
  
  \begin{tabular}{l | ccc ccc | ccc}
    \hline
    \multirow{2}{*}{Model} & \multicolumn{3}{c}{CBT-specific Skills} & \multicolumn{3}{c|}{General Counseling Skills} & \multirow{2}{*}{Turns} &  \multirow{2}{*}{Length} & \multirow{2}{*}{Modal.} \\
    \cline{2-7} 
    & Guided Disc. & Focus & Strategy & Underst. & Interp. Eff. & Collab. & \\
    \hline
    CACTUS & 
    3.99 $\mid$ \adjustbox{valign=m}{\fontsize{6pt}{6.5pt}\selectfont\shortstack{{\color{blue}+0.01}\\{\color{red}-0.03}}} & 
    3.96 $\mid$ \adjustbox{valign=m}{\fontsize{6pt}{6.5pt}\selectfont\shortstack{{\color{blue}-0.00}\\{\color{red}-0.07}}} & 
    4.20 $\mid$ \adjustbox{valign=m}{\fontsize{6pt}{6.5pt}\selectfont\shortstack{{\color{blue}+0.16}\\{\color{red}-0.30}}} & 
    4.60 $\mid$ \adjustbox{valign=m}{\fontsize{6pt}{6.5pt}\selectfont\shortstack{{\color{blue}-0.06}\\{\color{red}-0.14}}} & 
    5.99 $\mid$ \adjustbox{valign=m}{\fontsize{6pt}{6.5pt}\selectfont\shortstack{{\color{blue}+0.01}\\{\color{red}-0.02}}} & 
    4.50 $\mid$ \adjustbox{valign=m}{\fontsize{6pt}{6.5pt}\selectfont\shortstack{{\color{blue}+0.03}\\{\color{red}-0.40}}} & 
    23.99 & 69.87 & T \\
    
    MAGneT & 
    4.04 $\mid$ \adjustbox{valign=m}{\fontsize{6pt}{6.5pt}\selectfont\shortstack{{\color{blue}+0.23}\\{\color{red}-0.20}}} & 
    3.63 $\mid$ \adjustbox{valign=m}{\fontsize{6pt}{6.5pt}\selectfont\shortstack{{\color{blue}+0.26}\\{\color{red}-0.51}}} & 
    2.84 $\mid$ \adjustbox{valign=m}{\fontsize{6pt}{6.5pt}\selectfont\shortstack{{\color{blue}+0.61}\\{\color{red}-0.84}}} & 
    3.97 $\mid$ \adjustbox{valign=m}{\fontsize{6pt}{6.5pt}\selectfont\shortstack{{\color{blue}+0.03}\\{\color{red}-0.05}}} & 
    4.29 $\mid$ \adjustbox{valign=m}{\fontsize{6pt}{6.5pt}\selectfont\shortstack{{\color{blue}+0.45}\\{\color{red}-0.29}}} & 
    3.39 $\mid$ \adjustbox{valign=m}{\fontsize{6pt}{6.5pt}\selectfont\shortstack{{\color{blue}+0.55}\\{\color{red}-0.95}}} & 
    42.00 & 76.71 & T \\

    SQPsych & 
    4.42 $\mid$ \adjustbox{valign=m}{\fontsize{6pt}{6.5pt}\selectfont\shortstack{{\color{gray}+0.00}\\{\color{gray}-0.00}}} & 
    4.28 $\mid$ \adjustbox{valign=m}{\fontsize{6pt}{6.5pt}\selectfont\shortstack{{\color{gray}+0.00}\\{\color{gray}-0.00}}} & 
    4.50 $\mid$ \adjustbox{valign=m}{\fontsize{6pt}{6.5pt}\selectfont\shortstack{{\color{gray}+0.00}\\{\color{gray}-0.00}}} & 
    5.29 $\mid$ \adjustbox{valign=m}{\fontsize{6pt}{6.5pt}\selectfont\shortstack{{\color{gray}+0.00}\\{\color{gray}-0.00}}} & 
    \textbf{6.00} $\mid$ \adjustbox{valign=m}{\fontsize{6pt}{6.5pt}\selectfont\shortstack{{\color{gray}+0.00}\\{\color{gray}-0.00}}} & 
    \textbf{5.76} $\mid$ \adjustbox{valign=m}{\fontsize{6pt}{6.5pt}\selectfont\shortstack{{\color{gray}+0.00}\\{\color{gray}-0.00}}} & 
    30.61 & 39.61 & T \\

    MIRROR & 
    \underline{4.86} $\mid$ \adjustbox{valign=m}{\fontsize{6pt}{6.5pt}\selectfont\shortstack{{\color{blue}+0.86}\\{\color{red}-0.43}}} & 
    \underline{4.78} $\mid$ \adjustbox{valign=m}{\fontsize{6pt}{6.5pt}\selectfont\shortstack{{\color{blue}+1.06}\\{\color{red}-0.53}}} & 
    \underline{5.07} $\mid$ \adjustbox{valign=m}{\fontsize{6pt}{6.5pt}\selectfont\shortstack{{\color{blue}+0.93}\\{\color{red}-0.46}}} & 
    \underline{5.47} $\mid$ \adjustbox{valign=m}{\fontsize{6pt}{6.5pt}\selectfont\shortstack{{\color{blue}+0.27}\\{\color{red}-0.13}}} & 
    \textbf{6.00} $\mid$ \adjustbox{valign=m}{\fontsize{6pt}{6.5pt}\selectfont\shortstack{{\color{blue}+0.00}\\{\color{red}-0.00}}} & 
    \underline{5.71} $\mid$ \adjustbox{valign=m}{\fontsize{6pt}{6.5pt}\selectfont\shortstack{{\color{blue}+0.29}\\{\color{red}-0.15}}} & 
    20.25 & 26.00 & T/V \\
    
    \hdashline
    
    ODRA & 
    4.23 $\mid$ \adjustbox{valign=m}{\fontsize{6pt}{6.5pt}\selectfont\shortstack{{\color{blue}+1.45}\\{\color{red}-1.43}}} & 
    4.20 $\mid$ \adjustbox{valign=m}{\fontsize{6pt}{6.5pt}\selectfont\shortstack{{\color{blue}+1.78}\\{\color{red}-2.10}}} & 
    4.09 $\mid$ \adjustbox{valign=m}{\fontsize{6pt}{6.5pt}\selectfont\shortstack{{\color{blue}+1.91}\\{\color{red}-2.05}}} & 
    4.64 $\mid$ \adjustbox{valign=m}{\fontsize{6pt}{6.5pt}\selectfont\shortstack{{\color{blue}+1.35}\\{\color{red}-1.84}}} & 
    4.88 $\mid$ \adjustbox{valign=m}{\fontsize{6pt}{6.5pt}\selectfont\shortstack{{\color{blue}+1.12}\\{\color{red}-2.06}}} & 
    4.26 $\mid$ \adjustbox{valign=m}{\fontsize{6pt}{6.5pt}\selectfont\shortstack{{\color{blue}+1.71}\\{\color{red}-2.19}}} & 
    \underline{63.85} & 62.69 & T \\
    
    ODRA-NT & 
    4.01 $\mid$ \adjustbox{valign=m}{\fontsize{6pt}{6.5pt}\selectfont\shortstack{{\color{blue}+1.63}\\{\color{red}-1.61}}} & 
    3.96 $\mid$ \adjustbox{valign=m}{\fontsize{6pt}{6.5pt}\selectfont\shortstack{{\color{blue}+1.92}\\{\color{red}-1.90}}} & 
    3.72 $\mid$ \adjustbox{valign=m}{\fontsize{6pt}{6.5pt}\selectfont\shortstack{{\color{blue}+2.17}\\{\color{red}-1.72}}} & 
    4.19 $\mid$ \adjustbox{valign=m}{\fontsize{6pt}{6.5pt}\selectfont\shortstack{{\color{blue}+1.67}\\{\color{red}-1.79}}} & 
    4.44 $\mid$ \adjustbox{valign=m}{\fontsize{6pt}{6.5pt}\selectfont\shortstack{{\color{blue}+1.54}\\{\color{red}-2.23}}} & 
    3.92 $\mid$ \adjustbox{valign=m}{\fontsize{6pt}{6.5pt}\selectfont\shortstack{{\color{blue}+1.96}\\{\color{red}-1.92}}} & 
    \textbf{68.12} & 63.54 & T \\
    
    \bluebg ODRA-NR & 
    \bluebg \textbf{5.32} $\mid$ \adjustbox{valign=m}{\fontsize{6pt}{6.5pt}\selectfont\shortstack{{\color{blue}+0.55}\\{\color{red}-0.93}}} & 
    \bluebg \textbf{5.50} $\mid$ \adjustbox{valign=m}{\fontsize{6pt}{6.5pt}\selectfont\shortstack{{\color{blue}+0.48}\\{\color{red}-0.89}}} & 
    \bluebg \textbf{5.36} $\mid$ \adjustbox{valign=m}{\fontsize{6pt}{6.5pt}\selectfont\shortstack{{\color{blue}+0.63}\\{\color{red}-1.11}}} & 
    \bluebg \textbf{5.81} $\mid$ \adjustbox{valign=m}{\fontsize{6pt}{6.5pt}\selectfont\shortstack{{\color{blue}+0.17}\\{\color{red}-0.20}}} & 
    \bluebg \underline{5.92} $\mid$ \adjustbox{valign=m}{\fontsize{6pt}{6.5pt}\selectfont\shortstack{{\color{blue}+0.08}\\{\color{red}-0.08}}} & 
    \bluebg 5.44 $\mid$ \adjustbox{valign=m}{\fontsize{6pt}{6.5pt}\selectfont\shortstack{{\color{blue}+0.56}\\{\color{red}-0.95}}} & 
    57.03 & 62.14 & T \\
    \hline
  \end{tabular}
  \caption{\label{tab:ctrs}
    CTRS performance comparison of ODRA variants against existing baselines. Standard deviations are included to illustrate performance when isolating the evaluation of {\color{blue}positive} and {\color{red}negative} patients. ODRA variants include ODRA-NT (which deactivates TIBs classification) and ODRA-NR (which deactivates patient resistance). SQPsych lacks standard deviation values as it does not model patient attitudes. Metrics included are Guided Discovery, Focus, Strategy, Understanding, Interpersonal Efficiency, Collaboration, Average turns per session, Average therapist response length and Modalities. Possible modalities are text (\textbf{T}) and visual (\textbf{V}). ODRA-NR outperforms baselines in four out of six metrics, generating a significantly greater number of turns. We highlight the \textbf{best} and the \underline{second best} results.}
\end{table*}

\textbf{Models.} For session synthesis, we conduct an ablation study and select DeepSeek-V3.2 for its superior performance (see Appendix \ref{apx:model-ablation}). For fine-tuning models, see Section \ref{ssec:downstream-ft}.






\noindent \textbf{Baselines.} We evaluate against CACTUS \cite{lee-etal-2024-cactus} as the standard in CBT synthesis, MAGneT \cite{mandal2026magnet} and SQPsych \cite{vu2025roleplaying} for their multi-agent architectures, and MIRROR \cite{kim-etal-2025-mirror} for its specialized patient resistance modeling.


\noindent\textbf{Datasets.} We create a dataset of 150 ODRA synthetic CBT sessions, comprising a total of 9,577 turns and 18,496 reasoning traces. The conversations are generated from 150 patient intake forms from CACTUS evaluation dataset, encompassing 50 distinct patient profiles across three attitudinal variants. Because official baseline datasets utilize different patient profiles, we employ Sentence Transformers \cite{reimers2019sentencebert} to retrieve the most similar profiles for each case, achieving a profile similarity score of 0.75 (see Appendix \ref{apx:profiles-sim}). These datasets are used to compute global baseline metrics and to fine-tune models. To evaluate the fine-tuned models, we use an independent partition containing 150 extra patient intake forms from CACTUS evaluation dataset.


\noindent\textbf{Ablations.} We generate multiple ODRA dataset variants to evaluate the contribution of individual components. Standard dataset with all activated components is denoted as ODRA, and its variants are specified by the following suffixes: \textbf{-NR} indicates that \textit{resistance orchestrator} is deactivated, \textbf{-NT} indicates that \textit{TIBs classification} is deactivated, and \textbf{-T} indicates that \textit{reasoning traces} are included in the fine-tuning target (see Appendix \ref{apx:odra-variants}). 


\subsection{Automated Evaluation}
To assess ODRA's performance, we adopt a LLM-as-a-judge setup based on GPT-4o \cite{lee-etal-2024-cactus,wu2025axbench,ding2025selfexploring}. Appendix \ref{apx:metric-details} contains metrics definitions and evaluation prompts.

\noindent\textbf{CBT Counseling.} General and CBT skills are assessed using the Cognitive Therapy Rating Scale (CTRS) \cite{goldberg2020ctrs}, which contains six metrics rated on a 0-6 scale.\footnote{Guided Discovery, Focus, Strategy, Understanding, Interpersonal Effectiveness, and Collaboration} We also employ three metrics to evaluate the therapist resistance estimation: Accuracy ($\uparrow$), MAE ($\downarrow$) and RMSE ($\downarrow$).

\noindent\textbf{Behavioral Alignment.} To evaluate the efficacy of our Behavioral Profiler (see Section \ref{ssec:resistance}), we adopt the approach introduced in \citet{wu2025axbench}, with domain-specific modifications. Specifically, the LLM judge measures the following traits from patient utterances: Resistance Alignment, Contextual Alignment, and Realism. The judge scores each dialogue dimension independently on a 0-2 scale, aggregating results via harmonic mean to heavily penalize failure in any single dimension.

\noindent\textbf{Reasoning Traces.} To ensure the validity of our intermediate reasoning steps, we adapt the trace evaluation protocol from \citet{ding2025selfexploring}. Consistent with our steering evaluation, a LLM judge rates our reasoning traces across three metrics on a 0-2 scale, aggregating results via harmonic mean. The evaluation metrics are: Faithfulness, Logic Consistency, and Answer-explanation Alignment. Since our dataset contains a large number of traces, we implement a sampling strategy (see Appendix \ref{apx:metrics-details-defs}) to evaluate a highly representative subset.

\subsection{Expert Evaluation}
We conduct an expert evaluation involving three licensed psychologists who voluntarily participated in the study. First, experts blindly compare and rank groups of three conversations from CACTUS, MAGneT, and ODRA across thirteen different metrics spanning five categories: Therapeutic Skills, CBT Alignment, Fidelity, Specificity, and Safety. For this evaluation, we use a 1-3 ranking system where lower scores are better. Second, the experts evaluate ODRA reasoning traces along three axes on a 1–5 Likert scale: Contextual Adherence, Reasoning Adherence, and Clinical Assessment. We follow the sampling strategy described in Section \ref{apx:metrics-details-defs} to obtain these traces, but limited the scope to therapist ones, since our aim is to evaluate whether the traces contain accurate therapeutic information. We compute Krippendorff's $\alpha$ for the ranking task and \% agreement for the trace evaluation to ensure inter-annotator reliability.

A total of 50 samples are evaluated by at least two experts, resulting in 300 session evaluations, and 540 reasoning traces examined. Experts spent $\sim$1 hour reviewing each evaluation set. Details regarding annotator guidelines are in Appendix \ref{apx:experts-eval}.

\subsection{Downstream Fine-tuning}\label{ssec:downstream-ft}

To assess the capabilities of ODRA and baselines in adapting models to CBT counseling, we followed the approach in existing works \cite{lee-etal-2024-cactus,vu2025roleplaying} and fine-tune Llama-3-8B-Instruct. Additionally, we also trained Qwen-3.5-9B across all datasets to analyze the impact of training with ODRA reasoning traces. See Appendix \ref{apx:ft-details} for fine-tuning details.
\section{Results}

\subsection{CBT Counseling}\label{ssec:ctrs}

As shown in Table \ref{tab:ctrs}, ODRA-NR (our non-resistance variant) outperforms state-of-the-art methods, particularly across CBT skills. Conversely, the full ODRA exhibits lower performance than MIRROR, SQPsych and ODRA-NR, due to the challenging patient behaviors elicited when the resistance orchestrator is active. These behaviors mirror clinical reality where resistant patients may interrupt the CBT process (e.g., by rejecting the therapist's approach), increasing the difficulty of guiding the session, which ultimately results in lower counseling metrics. Furthermore, the ablation variant ODRA-NT underscores the role of the TIBs classification step, as its deactivation produces a performance drop compared to our full variant. ODRA results demonstrate that Focus and Strategy CBT skills are especially affected when dealing with resistant patients, a decline we attribute to the difficulty of maintaining CBT protocol adherence while addressing challenging behaviors. This behavioral complexity is reflected in the reported standard deviations. While deviations remain low for baseline methods and ODRA-NR, indicating therapeutic fluidity regardless of patient attitude, they are consistently high for ODRA and ODRA-NT, showing the distinct clinical frictions encountered when navigating positive and negative patients. Thus, for a fair evaluation against baselines under identical patient conditions, ODRA-NR serves as the appropriate benchmark. Due to the longer length of ODRA sessions, we analyzed the impact of turn volume on the final results and concluded that a higher number of turns is not correlated with better results (see Appendix \ref{apx:length}).


Regarding the patient resistance prediction, our therapist model achieved an Accuracy ($\uparrow$) of 0.80, a MAE ($\downarrow$) of 0.10, and a RMSE ($\downarrow$) of 0.13. See Appendix \ref{apx:resistance-prediction} for results details.



\subsection{Behavioral Alignment}\label{ssec:exp-steering}
Results for the behavioral alignment evaluation are shown in Table \ref{tab:steering}, where SQPsych is not included as it does not model patient attitudes. While all methods obtain comparable results for Context Alignment and Realism metrics, ODRA is the only framework that achieves consistently high scores within the Resistance Alignment dimension. The low scores achieved by the baselines and ODRA-NR show their misalignment with their ground-truth attitude, which may produce that negative patients produce overly compliant utterances. Conversely, the high scores yielded by ODRA shows its capability to mirror clinical reality, where negative patients exhibit clinical friction that impedes therapeutic progress. See Appendix \ref{apx:qualitative} for examples.

\begin{table}[htpb]
  \centering
  \small
  \setlength{\tabcolsep}{5.5pt}
  \renewcommand{\arraystretch}{1.25} 
  \begin{tabular}{lcccc | c}
    \hline
    Method & Res. Align. & Context & Realism & & HM \\
    \hline
    CACTUS          & 0.70 & \textbf{2.00} & 1.90 & & 0.69 \\
    MAGNET          & 1.19 & 1.97 & 1.80 & & 1.29 \\
    MIRROR          & 1.01 & 2.00 & 2.00 & & 1.01 \\
    \cdashline{1-6}
    \rowcolor{blue!10}
    ODRA     & \textbf{1.96} & 1.96 & \textbf{2.00} & & \textbf{1.95} \\
    ODRA-NR  & 1.18 & \textbf{2.00} & \textbf{2.00} & & 1.20 \\
    \hline
  \end{tabular}
  \caption{\label{tab:steering}
    Behavioral alignment comparison. Metrics include Resistance Alignment (\textbf{Res. Align.}), \textbf{Context} Alignment, Patient-like \textbf{Realism}, and Harmonic Mean (\textbf{HM}). We highlight the \textbf{best} results.
  }
\end{table}



\subsection{Reasoning Traces}\label{ssec:eval-rtraces}

The traces obtained a score of 1.84, 1.86, and 1.93 out of 2 for Faithfulness, Logic Consistency, and Answer-explanation Alignment, respectively. These values yield an aggregated harmonic mean of 1.83, confirming the high reliability of our CoT steps. See Appendix \ref{apx:qualitative} for traces examples.


\subsection{Expert Evaluation}

Table \ref{tab:expert_grouped_fidelity} reports the results for the blind ranking evaluation, where values represent the average from the items composing each category (see Appendix \ref{apx:experts-fg} for individual items results). Results obtained demonstrate the qualitative superiority of ODRA sessions, whose average rank is 1.09 with an inter-annotator agreement of $\alpha = 0.80$. Regarding the reasoning traces evaluated on a 1-5 Likert scale, experts rated our traces at 4.84 for Contextual Adherence, 4.86 for Reasoning Adherence, and 4.75 for Clinical Assessment. The experts conducted this evaluation with an average agreement of 96.05\%.

\begin{table}[htpb]
    \centering
    \small
    \setlength{\tabcolsep}{7.5pt}
    \renewcommand{\arraystretch}{1.25} 
    \begin{tabular}{lccccc}
        \hline
        Method & SK & CBT & SP & FD & SF (\%) \\
        \hline
        CACTUS & 2.83 & 2.61 & 2.81 & 2.80 & \textbf{0.00} \\
        MAGNET & 2.10 & 2.33 & 2.11 & 2.08 & 2.00 \\
        \rowcolor{blue!10}
        ODRA   & \textbf{1.07} & \textbf{1.07} & \textbf{1.08} & \textbf{1.12} & 1.00 \\
        \cdashline{1-6}
        Agreement & 0.88 & 0.71 & 0.75 & 0.84 & 0.98 \\
        \hline
    \end{tabular}
    \caption{\label{tab:expert_grouped_fidelity} Aggregated expert evaluation results, where lower is better. Metrics are: Therapeutic Skills (\textbf{SK}), CBT Alignment (\textbf{CBT}), Specificity (\textbf{SP}), Fidelity (\textbf{FD}), and Safety (\textbf{SF}). We highlight the \textbf{best} results.}
\end{table}

\subsection{Downstream Fine-tuning}\label{ssec:results-ft}


Table \ref{tab:ft} outlines the CTRS performance of Llama-3-8B-Instruct when fine-tuned on ODRA variants versus baselines across different patient configurations. For the non-resistant patient settings, ODRA-NR significantly outperforms baselines, achieving the largest gains in CBT skills. We evaluate ODRA-NR here to conduct a fair comparison under identical patient conditions. For the resistance setting, ODRA outperform both baselines and ODRA-NR, demonstrating that incorporating our patient resistance into dataset distributions produces more capable models.

While ODRA-NR largely surpasses baselines in non-resistant settings, MIRROR achieves competitive results under resistance. However, its short interactions, brief responses and narrow focus on cognitive reframing make it inappropriate for training CBT therapists (see Appendix \ref{apx:ft-llava}).

Finally, results for fine-tuning Qwen-3.5-9B across all datasets are detailed in Appendix \ref{apx:ft-qwen}. While ODRA variants achieve the highest scores, incorporating reasoning traces yields lower performance due to data volume and training limitations.

\begin{table}[htpb]
  \centering
  \fontsize{6.5pt}{8pt}\selectfont
  \setlength{\tabcolsep}{5pt} 
  \renewcommand{\arraystretch}{1.3}
  \newcommand{\supertiny}{\fontsize{4pt}{5pt}\selectfont\color{gray}}

  \begin{tabular}{l ccc ccc |cc}
    \hline
    Model & GD & FC & ST & UN & IE & CL & T & L \\
    \hline
    \multicolumn{9}{l}{\textit{w/o Resistance (DeepSeek Patient)}} \\
    \hline
    CACTUS    & 3.50 & 3.33 & 3.28 & 3.74 & 4.68 & 3.78 & 62.89 & 112.64 \\
    MAGNET    & 3.69 & 3.41 & 3.00 & 3.97 & 4.16 & 3.39 & 97.67 & 77.92 \\
    SQPsych   & 4.04 & 3.73 & \underline{3.90} & 4.31 & 5.00 & \underline{4.25} & 97.89 & 41.23 \\
    MIRROR    & \underline{4.54} & \underline{4.11} & 3.78 & \underline{5.05} & \underline{5.24} & 4.23 & 83.65 & 23.57 \\
    \rowcolor{blue!10}
    ODRA-NR   & \textbf{5.11} & \textbf{5.16} & \textbf{4.96} & \textbf{5.71} & \textbf{5.91} & \textbf{4.99} & 84.83 & 62.27 \\
    \hline
       
    \multicolumn{9}{l}{\textit{w/o Resistance (GPT Patient)}} \\
    \hline
    CACTUS    & 3.96 & 3.88 & 3.92 & 4.32 & 5.96 & 4.34 & 48.57 & 101.88 \\
    MAGNET    & \underline{4.81} & 4.01 & \underline{4.40} & 5.40 & 5.78 & \underline{5.45} & 50.00 & 82.28 \\
    SQPsych   & 4.79 & \underline{4.20} & 4.32 & 4.71 & \underline{5.97} & \textbf{5.87}  & 48.49 & 33.53 \\
    MIRROR    & 4.63 & 4.19 & 3.93 & \underline{5.44} & 5.95 & 5.33 & 42.97 & 24.81 \\
    \rowcolor{blue!10}
    ODRA-NR   & \textbf{5.48} & \textbf{5.55} & \textbf{5.24} & \textbf{5.71} & \textbf{6.00} & 5.37 & 47.97 & 55.06 \\
    \hline
       
    \multicolumn{9}{l}{\textit{w/ Resistance (DeepSeek Patient)}} \\
    \hline
    MAGNET    & 3.56 & 3.20 & 3.18 & 3.62 & 4.09 & 3.56 & 85.68 & 77.66 \\
    CACTUS    & 3.00 & 2.88 & 2.94 & 3.08 & 3.98 & 3.29 & 75.65 & 111.01 \\
    SQPsych   & 3.51 & 3.33 & 3.30 & 3.56 & 4.13 & 3.79 & 80.24 & 40.35\\
    MIRROR    & 3.77 & 3.68 & 3.54 & 4.15 & \underline{4.55} & \underline{4.13} & 56.97 & 24.47 \\
    ODRA-NR   & \underline{4.08} & \underline{3.95} & \underline{3.72} & \underline{4.29} & \underline{4.55} & 4.12 & 71.28 & 68.83 \\
    \rowcolor{blue!10}
    ODRA      & \textbf{4.19} & \textbf{4.07} & \textbf{3.78} & \textbf{4.35} & \textbf{4.71} & \textbf{4.20} & 69.91 & 63.85 \\
    \hline
  \end{tabular}
  \caption{\label{tab:ft} Comparison between Llama-3 fine-tuned on different state-of-the-art datasets. Metrics included are: Guided Discovery (\textbf{GD}), Focus (\textbf{FC}), Strategy (\textbf{ST}), Understanding (\textbf{UN}), Interpersonal Effectiveness (\textbf{IE}), Collaboration (\textbf{CL}), Avg. Turns (\textbf{T}), and Therapist Utterance Avg. Length (\textbf{L}). We highlight the \textbf{best} and the \underline{second best} CTRS metric results for each section.}
\end{table}

\section{Conclusions}
In this work, we introduced ODRA, a novel method for synthesizing therapy sessions. We developed a Chain-Of-Thought objective-driven framework that strictly adheres to the foundational Cognitive Behavioral Therapy protocol. Furthermore, we implemented a Resistance Orchestrator that manages patient resistance dynamics, effectively mitigating LLM sycophancy through fine-grained prompt steering techniques. Our automated and expert assessments demonstrate the significant superiority of ODRA in therapeutic capabilities and behavioral alignment. Ultimately, models fine-tuned on ODRA datasets exhibited superior downstream clinical performance, positioning ODRA as a significant advancement toward synthetic data generation for mental health applications.



\section*{Limitations}

\noindent\textbf{Longitudinal Paradigm and Session Length.} Our framework adapts the longitudinal nature of CBT into a single-session paradigm. Consequently, this prevents the simulation of some structural components defined in foundational CBT guidelines, such as reviewing homework or summarizing insights from previous sessions. This constraint limits modeling patient progression over time and adjusting therapeutic strategies accordingly. Additionally, while ODRA generates significantly longer interactions than existing baselines, the average turn count remains below real-world clinical sessions, which typically span an hour.

\smallskip
\noindent\textbf{Intake Forms as Patient Inputs.} While patient intake forms accurately simulate the initial information available during a first session, transitioning to longitudinal modeling requires more sophisticated structures that enable dynamic representation of patient internal states, such as evolving cognitive models. Future work should explore how to integrate more complex patient representations into the workflow of our framework.

\smallskip
\noindent\textbf{Computational Scaling Costs.} Our multi-stage Chain-of-Thought strategy relies on sequential prompt steps to ensure CBT protocol adherence and maximize utterance quality. While downstream LLMs can be fine-tuned on small-scale datasets for domain adaptation, scaling synthesis to meet the data volume needed to train highly capable therapist models would require a substantial computational budget. Future work must focus on token efficiency to facilitate large-scale dataset synthesis.

\smallskip
\noindent\textbf{Mental Health Models Grounding.} Our framework currently leverages general-purpose foundational LLMs that lack specialized pre-training on clinical psychology. Integrating models pre-trained on authentic counseling interactions could further elevate session realism by capturing therapeutic nuances that remain difficult to simulate zero-shot, such as detecting implicit patient ambivalence or delivering micro-interventions at appropriate clinical moments.

\smallskip
\noindent\textbf{Therapist Resistance Handling.} In this work, we primarily focus on modeling resistance from the patient side. However, developing specialized therapist modules is essential to mitigate the performance drops observed when confronting highly resistant patient profiles (see Section \ref{ssec:results-ft}). Integrating these components paves the way for the development of frameworks capable of generating a broader spectrum of clinical scenarios.

\smallskip
\noindent\textbf{LLM-as-a-judge.} Despite GPT-based LLM-as-a-judge setups are a standard evaluation protocol, they introduce biases, such as the length bias shown in our downstream experiments (see Section \ref{ssec:results-ft}). Furthermore, relying on proprietary models constraints the capabilities of research labs with limited resources, making the development of cost-effective alternatives essential for fostering synthetic CBT research. In this work, we complemented the automated evaluation with an expert assessment to validate the quality of our synthetic sessions.


\section*{Ethics Statement}

\textbf{Privacy Considerations.} Although this framework can be seamlessly adapted to use real-world patient profiles, all the experiments conducted in this study were performed exclusively using data extracted from CACTUS dataset \cite{lee-etal-2024-cactus}, which uses PATTERNREFRAME \cite{maddela-etal-2023-training} as its seed dataset. This dataset does not contain medical records, instead it relies on crowdsourcing where participants were instructed to write sentences aligned with a specific persona profile and a negative thought pattern. 

\smallskip
\noindent\textbf{Safety.} We conducted an expert evaluation where psychologists assessed the safety of generated sessions. While results indicate that a large majority of sessions are safe, a comprehensive analysis would be required to identify edge cases where the framework might output invalid or harmful content.

\smallskip
\noindent\textbf{Clinical Deployment.} Our work aims at advancing the research field of synthetic counseling sessions, but more comprehensive evaluations are a strict prerequisite before utilizing data synthesized by this framework in real-world clinical deployments. Models trained on this data should be carefully evaluated through clinical trials to assess their therapeutic safety, correctness and behavioral alignment, especially conversational models intended to interact with real clients. While this work actively fosters the development of LLM-based therapists, we emphasize the importance of their usage under the supervision of certified mental health professional within clinical environments.




\bibliography{custom}

\clearpage
\appendix
\section{Cognitive Behavioral Therapy Data}\label{apx:beck-theory}
This section compiles all the theoretical data regarding the foundational CBT protocol we used as basis to construct the ODRA framework.

\subsection{Core Beliefs}
CBT guidelines defines a list of different core belief types that can be used to complete a cognitive model.\footnote{Extracted from \url{https://learn.beckinstitute.org/s/product/cbt-worksheet-packet/01t4M000004NMqnQAG}} We will specifically use unhelpful core beliefs which expands to three types: \textit{Helpless}, \textit{Unlovable}, and \textit{Worthless} core beliefs.

\subsection{Cognitive Distortions}

Concerning cognitive distortions, we use a subset from the original taxonomy proposed in CBT guidelines to ensure alignment with the categories present in the CACTUS evaluation set, which serves as our baseline. To achieve this, we obtained a list of unique cognitive distortions appearing in CACTUS, and we filtered out from the original CBT list the distortions not included. The final list of cognitive distortions alongside their definitions\footnote{Definitions extracted them from the Depression Information Sheet: \textit{Unhelpful Thinking Styles}, that can be accessed via \url{https://www.cci.health.wa.gov.au/resources/looking-after-yourself/depression}} are available next:

\begin{itemize}
    \item \textbf{Mental filtering.} This thinking styles involves a 'filtering in' and 'filtering out' process - a sort of 'tunnel vision,' focusing on only one part of a situation and ignoring the rest. Usually this means looking at the negative parts of a situation and forgetting the positive parts.
    \item \textbf{Jumping to conclusions: fortune-telling.} We jump to conclusions when we assume that we know what someone else is thinking.
    \item \textbf{Jumping to conclusions: mind-reading.} We jump to conclusions when we assume that we know what is going to happen in the future.
    \item \textbf{Personalization}: This involves blaming yourself for everything that goes wrong or could go wrong, even when you may only be partly responsible or not responsible at all. You might be taking 100\% responsibility for the occurrence of external events.
    \item \textbf{Catastrophizing.} This occurs when we 'blow things out of proportion', and we view the situation as terrible, awful, dreadful, and horrible, even though the reality is that the problem itself is quite small.
    \item \textbf{Black-and-white or polarized thinking / all or nothing thinking}: This thinking style involves seeing only one extreme or the other. You are either wrong or right, good or bad and so on. There are no in-betweens or shades of gray.
    \item \textbf{Should statements.} Sometimes by saying 'I should...' or 'I must...' you can put unreasonable demands or pressure on yourself and others. Although these statements are not always unhelpful, they can sometimes create unrealistic expectations.
    \item \textbf{Overgeneralization.} When we overgeneralise, we take one instance in the past or present, and impose it on all current or future situations. If we say 'You always...' or 'Everyone...', or 'I never...' then we are probably overgeneralising.
    \item \textbf{Labeling and mislabeling.} We label ourselves and others when we make global statements based on behaviour in specific situations. We might use this label even though there are many more examples that aren't consistent with that label.
    \item \textbf{Discounting the positive.} In this thinking style, you magnify the positive attributes of other people and minimise your own positive attributes. It's as though you're explaining away your own positive characteristics.
\end{itemize}

\subsection{CBT Techniques}\label{apx:cbt-techniques}
During the CBT Work stage, different techniques are considered to address cognitive distortions. For this work, we used the same set of techniques that were used in CACTUS, which are listed with their definitions below:

\begin{itemize}
    \item \textbf{Efficiency Evaluation}: Assists individuals in evaluating the usefulness of their thoughts or beliefs, analyzing how practical or detrimental they are in real-life situations.
    \item \textbf{Pie Chart Technique}: Used for individuals experiencing excessive self-blame or responsibility, visually representing the contribution of various factors to a specific event or outcome.
    \item \textbf{Alternative Perspective}: Involves asking clients how others might think in similar situations, encouraging consideration of different interpretations.
    \item \textbf{Decatastrophizing}: Aims to reduce the tendency to imagine the worst-case scenario by evaluating the actual likelihood of the feared outcome and preparing for coping strategies.
    \item \textbf{Pros and Cons Analysis}: Analyzes the advantages and disadvantages of specific thoughts or beliefs, fostering a more balanced evaluation.
    \item \textbf{Evidence-Based Questioning}: Guides clients to find evidence supporting or contradicting their thoughts, promoting a more evidence-based approach to thinking.
    \item \textbf{Reality Testing}: Explores how well clients’ thoughts align with reality, helping them distinguish between thoughts and actual experiences.
    \item \textbf{Continuum Technique}: Positions clients’ experiences between two extreme situations, encouraging a more nuanced evaluation of situations.
    \item \textbf{Changing Rules to Wishes}: Replaces strict rules or arbitrary attitudes with realistic hopes or wishes.
    \item \textbf{Behavior Experiment}: Involves trying out new behaviors in specific situations to challenge and modify negative beliefs.
    \item \textbf{Problem-Solving Skills Training}: Learning systematic methods for resolving problem situations. This involves identifying problems, finding possible solutions, and implementing those solutions.
    \item \textbf{Systematic Exposure}: Gradual exposure to situations that cause fear or anxiety, allowing individuals to experience anxiety while learning how to manage it.
    \item \textbf{Cognitive Restructuring}: The process of identifying, evaluating, and responding to adaptive thoughts and beliefs using Socratic questioning and evidence-gathering.
    \item \textbf{Role Playing}: A session-based simulation of real-life interactions to practice new skills or gain a different perspective on a belief.
    \item \textbf{Activity Scheduling}: Planning and engaging in activities that are enjoyable or provide a sense of accomplishment to counteract negative thoughts and improve mood.
\end{itemize}

\section{Single-session CBT Adaptation}\label{apx:beck-adaptations}
While most sessions follow the same structure, the first session (often called Evaluation session), is completely different, as it is entirely focused on gathering information and determining whether CBT is appropriate for the patient. To adapt this longitudinal therapeutic process into a single session paradigm, we introduced the following adaptations to the foundational CBT protocol:

\begin{itemize}
    \item \textbf{Opening.} In order to simulate the initial problem identification characteristic of an Evaluation session, we extended the Opening stage to focus the therapist in extracting fundamental clinical data, which is a required for identifying cognitive models in the second stage. While a standard opening incorporates both a Mood Check and a Session Bridge (summarizing insights from the prior session), we omitted the session bridge due to the single-session constraint.
    \item \textbf{Homework Review.} Since CBT is a task-oriented protocol, therapists typically begin sessions assessing the patient progress with the tasks assigned in the previous session. Similarly to the Session Bridge, this component is omitted as no prior session exist.
    \item \textbf{Agenda Setting.} A therapeutic process frequently involves navigating through different patient concerns. Conversely, the patient intake forms we use as generation seeds provides only one problem per profile. Therefore, we omitted the Agenda Setting as the therapeutic objective is predefined by the input data.
    \item \textbf{Cognitive Conceptualization.} While not defined as a standalone stage in foundational CBT guidelines, we introduce this phase to replicate the initial discovery of cognitive models that a therapist conducts during the Evaluation session. In real-world CBT, conceptualizing cognitive models is a continuous, iterative task performed by both patient and therapist during the therapeutic process.
\end{itemize}
\section{Detailed CBT Aligner Implementation}\label{apx:cbtgrounder-details}
This section provides the specific implementation details of the the different stages composing the CBT Aligner component. First, we introduce the different reasoning steps composing the therapist Chain-of-Thought process across each session stage. Then, we present the controllability thresholds established to regulate session dynamics.

\subsection{Reasoning Traces}
\begin{figure}[t]
  \includegraphics[width=\columnwidth]{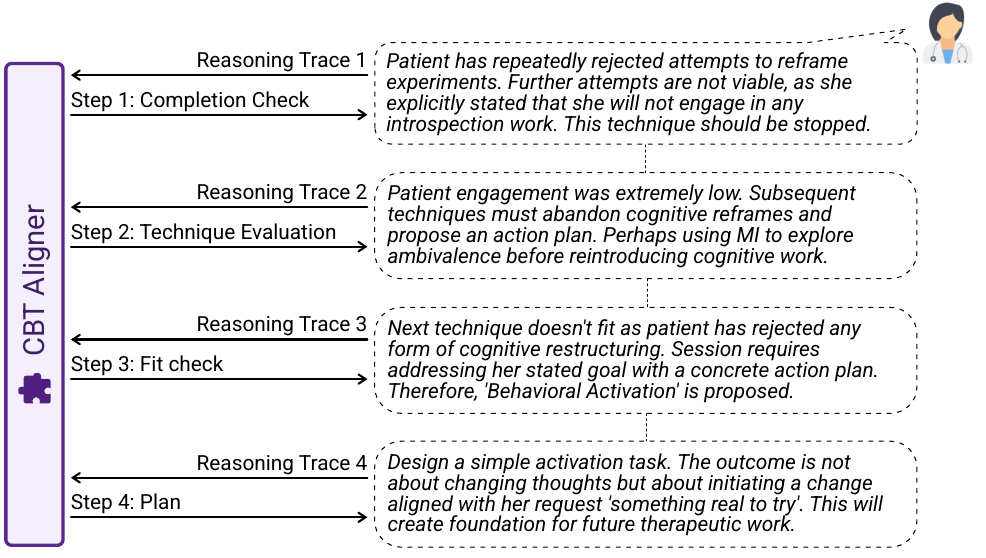}
  \caption{Execution example of the stage-specific reasoning steps generated in one Therapist turn when interacting with the CBT Aligner.}
  \label{fig:cbt-aligner}
\end{figure}

In this section, we introduce the different reasoning traces that constitute our framework, organized by stage. For each of the reasoning steps the therapist is instructed to provide a rationale alongside the expected output. Furthermore, Figure \ref{fig:cbt-aligner} provides a therapist chain-of-thought execution example. \\

\noindent\textbf{Global Traces.} Reasoning traces under this group are executed in each therapist turn:
\begin{itemize}
    \item \textbf{Resistance Prediction:} Estimates patient resistance.
    \item \textbf{Therapy-Interfering Behavior (TIB) Check:} Analyzes patient utterances to identify TIBs.
\end{itemize}

\noindent\textbf{TIBs Stage.} This stage is triggered for one turn when a TIB is detected by the global reasoning trace \textit{TIB Check}.
\begin{itemize}
    \item \textbf{TIB Plan Generation:} Formulates the plan to address the identified resistant behavior.
\end{itemize}

\noindent\textbf{Opening}
\begin{itemize}
    \item \textbf{Information Update:} Updates the recorded patient mood and presenting problem.
    \item \textbf{Readiness Check:} Validates whether enough information has been gathered to transition to the next stage.
\end{itemize}

\noindent\textbf{Cognitive Conceptualization}
\begin{itemize}
    \item \textbf{Bootstrap:} Constructs the initial cognitive model from the information gathered during the Opening stage.
    \item \textbf{Gap Planning:} Creates a plan to elicit missing cognitive components using Socratic questioning.
    \item \textbf{Gap Response Review:} Analyzes the patient's reply to verify if the targeted gap was successfully addressed, extracting any newly exposed cognitive items.
    \item \textbf{Refinement Planning:} Creates a plan to refine underexplored cognitive model items.
    \item \textbf{Distortion Probability Estimation:} Assigns the most likely cognitive distortion to the currently explored cognitive model.
    \item \textbf{Core Belief Probability Estimation:} Assigns the most likely core belief to the currently explored cognitive model.
\end{itemize}

\noindent\textbf{CBT Technique}
\begin{itemize}
    \item \textbf{Completion Check:} Validates whether the current technique can be completed.
    \item \textbf{Technique Selection:} Selects the most appropriate CBT technique based on the cognitive distortions identified.
    \item \textbf{Fit Check:} Assesses if the next technique to be applied is still feasible for the current session status.
    \item \textbf{Plan:} Outlines a plan detailing how the technique will be executed, and the expected outcomes.
    \item \textbf{Evaluation:} Conducts an evaluation of the completed technique, including outcomes and limitations encountered.
    \item \textbf{Apply More Techniques Check:} Determines the suitability of applying more techniques.
\end{itemize}

\noindent\textbf{Homework}
\begin{itemize}
    \item \textbf{Task Proposal:} Creates the list of tasks proposed to be executed by the patient.
    \item \textbf{Agreement Check:} Validates whether the patient has agreed to the proposed tasks.
\end{itemize}

\noindent\textbf{Closing}
\begin{itemize}
    \item \textbf{Summary Generation:} Creates a summary of the session for sharing with the patient.
    \item \textbf{Next Session Plan Generation:} Creates a next session plan for sharing with the patient.
    \item \textbf{Plan Agreement Check:} Validates whether the patient has agreed to the proposed next session plan.
\end{itemize}

\subsection{Controllability}
To ensure a balance between therapeutic load and session length, we utilized controllability thresholds in each stage. These thresholds were validated by the licensed experts involved in this work:

\begin{itemize}
    \item \textbf{Opening.} At each turn, the therapist model is prompted to validate if enough mood and problem information has been collected to proceed. If not, the therapist continues extracting information. To prevent conversational deadlock, we impose a hard threshold of six turns for this gathering phase. This threshold aligns with the initial engagement boundaries established by \cite{park2019designing}.
    \item \textbf{Cognitive Conceptualization.} When the therapist has already identified three cognitive models, it is prompted to validate whether the extracted models are enough to understand the patient difficulties. If not, the therapist initializes the discovery of a new one. This step addresses scenarios where initial models focus on redundant situations, that prevents the analysis of patient behaviors across different domains. To maximize contextual diversity, the model is explicitly prompted to explore heterogeneous situations. We set a maximum cap of five cognitive models for this phase. Furthermore, to prevent the therapist quickly completing cognitive models without enough therapeutic discovery, we enforce a minimum duration of three turns per cognitive model. 
    \item \textbf{CBT Work.} A minimum of three CBT techniques are utilized in this stage. Following the third technique, the therapist assesses whether sufficient therapeutic progress has been achieved. If further intervention is required, it selects an additional technique. We qualitatively observed that some techniques—such as behavioral experiments—can be overly exploratory, so we established a maximum of twelve turns per technique. Once this threshold is reached, the therapist smoothly ends the technique to begin a new one or to transition to the next stage. Similarly to cognitive models, a maximum cap of five CBT techniques is established.
    \item \textbf{Homework.} Because CBT is based on therapist-patient collaboration, the homework assignments must be mutually agreed upon. To do so, the therapist iteratively propose tasks until an agreement with the patient is reached. When interacting with negative patients, agreement can be difficult to achieve, so a maximum cap of five turns is established to achieve an agreement. During these iterations, the model dynamically generate less demanding variations from the initial set of tasks. If consensus is not reached within five turns, the framework continues with the next stage.
    \item \textbf{Closing.} During this final stage, the therapist attempts to reach a consensus regarding the next session plan. Consistently with other stages, a maximum of five turns is established to reach the agreement. If an agreement cannot be formalized, the therapist bypasses the loop to execute the next part of the stage.
\end{itemize}

The conditions and thresholds established not only ensure the quality of the generated sessions, but also help ODRA fine-tuned models to internalize expert therapeutic decision-making.

\section{Evaluation Profiles Similarity}\label{apx:profiles-sim}

The official dataset releases for some of the baselines utilized in our experiments employ different input patient profiles. To conduct a rigorous comparison, we use semantic similarity to match our evaluation profiles with those of the baseline methods. First, we compute an embedding for each target profile using a Sentence Transformer model\footnote{We utilize the \texttt{all-MiniLM-L6-v2} model.} \cite{reimers2019sentencebert}. We then compute the cosine similarity between each profile embedding in our evaluation dataset and all profile embeddings within a given baseline, extracting the closest matching baseline profile.

Although our evaluation dataset comprises 150 sessions, it is built upon 50 unique patient profiles ($50 \text{ profiles} \times 3 \text{ attitudes}$). Because baselines do not utilize our exact resistance labels, we cannot expand their datasets to 150 samples simply by varying the attitude field. To address this, we first retrieve the 50 most semantically similar profiles for each baseline, and subsequently expand them to 150 variants using baseline-specific strategies:

\noindent\textbf{MIRROR.} This framework models resistance using a specific resistance label or a \emph{no resistance} designation. For each matched baseline profile, we include its \emph{no resistance} variant and sample one of its existing resistant configurations. Because the majority of MIRROR profiles contain only a single resistant variant, we retrieve the second most semantically similar baseline profile to serve as the third variant, utilizing it in its resistant configuration. This procedure yielded an average profile similarity score of 0.75.

\noindent\textbf{CACTUS \& MAGneT.} These frameworks employs the exact same patient profiles and attitude labels as ours. Therefore, a direct profile matching without semantic similarity is performed.

\noindent\textbf{SQPsych.} This framework does not include any client profile information, which prevents similarity matching. For this reason, we created an evaluation set for this method by randomly sampling 150 dialogues from the initial training dataset. Specifically, we used the SQPsychConv\_qwen-2.5 dataset variant, which achieved the highest score in the human evaluation of the official paper.
\section{Evaluation Metrics Details}\label{apx:metric-details}
\begin{table*}[t!]
  \centering
  \small 
  \setlength{\tabcolsep}{6pt} 
  \renewcommand{\arraystretch}{1.5} 
  
  \newcommand{\bluebg}{\cellcolor{blue!10}}
  
  \begin{tabular}{l | ccc ccc}
    \hline
    \multirow{2}{*}{Model} & \multicolumn{3}{c}{CBT-specific Skills} & \multicolumn{3}{c}{General Counseling Skills} \\
    \cline{2-7} 
    & Guided Disc. ($\uparrow$) & Focus ($\uparrow$) & Strategy ($\uparrow$) & Underst. ($\uparrow$) & Interp. Eff. ($\uparrow$) & Collab. ($\uparrow$) \\
    \hline
    $[-0.05, +0.10]$ & 
    4.01 & 
    \underline{4.06} & 
    3.93 & 
    4.60 & 
    4.83 & 
    4.12 \\
    
    \bluebg $[-0.10, +0.15]$ & 
    \bluebg \underline{5.08} & 
    \bluebg \textbf{5.29} & 
    \bluebg \textbf{5.14} & 
    \bluebg \textbf{5.90} & 
    \bluebg \textbf{5.90} & 
    \bluebg \textbf{5.32} \\
    
    $[-0.20, +0.30]$ & 
    \textbf{5.09} & 
    5.14 & 
    \underline{5.09} & 
    \underline{5.49} & 
    \underline{5.69} & 
    \underline{5.22} \\
    \hline
  \end{tabular}
  \caption{\label{tab:hyperparameter-tuning-bounds} Hyperparameter tuning configurations exploring the impact of different resistance update boundaries in global session quality. A fixed homeostatic reversion rate of 0.15 was maintain across this experiment. We highlight the \textbf{best} and the \underline{second best} results.}
\end{table*}

\begin{table*}[t!]
  \centering
  \small 
  \setlength{\tabcolsep}{8pt} 
  \renewcommand{\arraystretch}{1.5} 
  
  \begin{tabular}{l | ccc ccc}
    \hline
    \multirow{2}{*}{Model} & \multicolumn{3}{c}{CBT-specific Skills} & \multicolumn{3}{c}{General Counseling Skills} \\
    \cline{2-7} 
    & Guided Disc. & Focus & Strategy & Underst. & Interp. Eff. & Collab. \\
    \hline
    0.05 & 
    \underline{4.81} & 
    \underline{4.85} & 
    \underline{4.70} & 
    \underline{5.28} & 
    \underline{5.44} & 
    \underline{4.88} \\
    
    \rowcolor{blue!10} 0.15 & 
    \textbf{5.08} & 
    \textbf{5.29} & 
    \textbf{5.14} & 
    \textbf{5.90} & 
    \textbf{5.90} & 
    \textbf{5.32} \\
    
    0.30 & 
    4.11 & 
    4.13 & 4.01 & 
    4.85 & 
    4.98 & 
    4.37 \\
    \hline
  \end{tabular}
  \caption{\label{tab:hyperparameter-tuning-er} Hyperparameter tuning configurations evaluating impact of homeostatic reversion rates under fixed resistance update boundaries (Bounds = $[-0.10, +0.15]$). We highlight the \textbf{best} and the \underline{second best} results.}
\end{table*}

In this section we provide further details about the different automated metrics we used throughout the evaluation of our framework. Furthermore, in this section we provide the LLM-as-a-judge evaluation prompts.

\subsection{Metrics Description}\label{apx:metrics-details-defs}
Next we provide details about the specific items assessed in each of the evaluations performs. We used a LLM-as-a-judge setup establishing the number of completions to three in order to ensure evaluation robustness. The final score is the average from the three evaluation completions.

\textbf{CBT Counseling.} To assess therapist counseling skills we used CounselingEval \cite{lee-etal-2024-cactus}, which is grounded in the CTRS. This is a tool to measure the way on which therapists deliver CBT. CounselingEval established six evaluation metrics covering CBT-specific and general counseling skills. The CBT-specific items are defined as follows:
\begin{itemize}
    \item \textbf{Guided Discovery:} Measures how well the therapist used the Guided Discovery technique to explore problems and help patient draw their own conclusions.
    \item \textbf{Focus:} Measures how well the therapist focused on key thoughts, assumptions and behaviors that were most relevant to the problem.
    \item \textbf{Strategy:} Measures how well the therapist followed a consistent strategy for change that seemed very promising and incorporated the most appropriate cognitive behavioral therapy techniques.
\end{itemize}
On the other side, the general counseling items are defined as follows:
\begin{itemize}
    \item \textbf{Understanding:} Measures how well the therapist seemed to understand the patient’s “internal reality” thoroughly and was adept at communicating this understanding through appropriate verbal and non-verbal responses to the patient
    \item \textbf{Interpersonal Effectiveness:} Measures how well the therapist displayed optimal levels of warmth, concern, confidence, genuineness, and professionalism.
    \item \textbf{Collaboration:} Measures how well the therapist encouraged patient as much as possible to take an active role during the session, fostering a collaborative environment.
\end{itemize}

\noindent\textbf{Therapist's Resistance Prediction.} To evaluate the therapist capability in predicting patient resistance we use Accuracy ($\uparrow$), MAE ($\downarrow$), and RMSE ($\downarrow$). While Accuracy measures how effectively the therapist infers the correct categorical patient attitude interval, MAE and RMSE quantify the magnitude of the prediction error with respect to the continuous resistance value. For computing the classification accuracy we mapped the therapist inferred value to its corresponding attitudinal interval and compared it against the ground-truth patient attitude interval for that turn. In terms of MAE and RMSE, we directly compared the inferred value against the patient's actual resistance value.

\noindent\textbf{Behavioral Alignment.} To evaluate the effectiveness of our behavioral profiler mechanism, we use the next metrics:
\begin{itemize}
    \item Resistance Alignment: Evaluates whether the patient's resistance $r$ is accurately reflected in the dialogue.
    \item Contextual Alignment: Evaluates whether the patient responses are related to the immediately preceding therapist utterances.
    \item Realism: Evaluates whether the patient utterances sound realistic and plausible for a therapy dialogue.
\end{itemize}

\begin{table*}
  \centering
  \small 
  \setlength{\tabcolsep}{6pt} 
  \renewcommand{\arraystretch}{1.5} 
  
  \begin{tabular}{l | ccc ccc | cc}
    \hline
    \multirow{2}{*}{Model} & \multicolumn{3}{c}{CBT-specific Skills} & \multicolumn{3}{c|}{General Counseling Skills} & \multirow{2}{*}{Turns} & \multirow{2}{*}{Words/Turn} \\
    \cline{2-7} 
    & Guided Disc. & Focus & Strategy & Underst. & Interp. Eff. & Collab. & & \\
    \hline
    \rowcolor{blue!10} DeepSeek-V3.2 & 
    \textbf{5.32} & 
    \textbf{5.50} & 
    \textbf{5.36} & 
    \underline{5.81} & 
    \textbf{5.92} & 
    \textbf{5.44} & 
    57.03 & 
    62.14 \\
    
    Qwen-3.5-122B & 
    \underline{5.08} & 
    \underline{5.29} & 
    \underline{5.14} & 
    \textbf{5.90} & 
    \underline{5.90} & 
    \underline{5.32} & 
    71.15 & 
    120.97 \\
    
    Llama-3-70B & 
    4.14 & 
    3.96 & 
    3.60 & 
    4.20 & 
    \underline{4.98} & 
    3.67 & 
    85.93 & 
    95.39 \\
    \hline
  \end{tabular}
  \caption{\label{tab:model-ablation} Therapist model ablation performance comparison across different LLM architectures on CBT and general counseling skills. We highlight the \textbf{best} and the \underline{second best} results.}
\end{table*}

For the rigorous evaluation of the resistance alignment metric, we implemented a validation constraint on which the judge rates each session against the three attitude definitions. If any non-target attitude receives a higher score than the expected target attitude, the Resistance Alignment score is zeroed out. This effectively penalizes dialogues where the judge confuses patient attitude, which is an indicator of behavioral misalignment. Since MIRROR provides resistance types instead of attitude labels, we map the \emph{no resistance} label to the \emph{positive} attitude, and any resistance label to the \emph{negative} attitude. Furthermore, as ODRA patient resistance is dynamic, we apply a resistance-based dialogue segmentation to separately evaluate dialogue segments with different patient attitude intervals. Then, we compute session-based averages before calculating the final results to avoid over-weighting more fragmented sessions. Finally, we aggregate results via harmonic mean to heavily penalize failure in any single dimension. This harmonic mean is calculated as the average of the harmonic means for each session, an approach taken to include in the global scores local metric failures, rather than globally measuring the performance of each metric. We define the global harmonic mean as follows:

\begin{equation}
    \text{HM} = \frac{1}{N} \sum_{i=1}^{N} \frac{M}{\sum_{j=1}^{M} \frac{1}{x_{i,j}}}
\end{equation}

\noindent where $M$ is the number of metrics and $N$ the number of sessions.

\noindent\textbf{Reasoning Traces.} To ensure the validity of our reasoning traces, we use the next metrics:
\begin{itemize}
    \item Faithfulness: Evaluates whether the reasoning steps are grounded in the previous conversation context.
    \item Logic Consistency: Evaluates whether the reasoning steps are logical and coherent between them.
    \item Answer-explanation Alignment: Evaluates whether the final utterance is aligned with the previously generated reasoning steps.
\end{itemize}

Due to the large volume of reasoning traces generated across the evaluation set, we utilized a stage-based sampling strategy. For each session, we randomly selected one therapist turn per each therapeutic stage and the subsequent patient intervention. Since traces within the same stage maintain a consistent structure and therapeutic reasoning, this approach allows us to evaluate a representative subset that reflects overall quality. Using this method, we evaluated a total of 821 turns, comprising 5268 reasoning traces (3676 therapist-side + 1592 patient-side).
\section{Resistance Prediction}\label{apx:resistance-prediction}

In each turn, the therapist estimates the ground-truth patient resistance to adapt its next utterance to the new resistance level and conduct the CBT session more effectively. We evaluate the therapist model's capability to infer this resistance using Accuracy ($\uparrow$), MAE ($\downarrow$), and RMSE ($\downarrow$), yielding scores of 0.80, 0.10, and 0.13, respectively. The high accuracy shows that the therapist is able to correctly infer the patient attitudinal interval, enabling it to adapt to the patient global stance toward therapy (Positive, Neutral, or Negative). Furthermore, the low MAE and RMSE values indicate minimal differences between the predicted and ground-truth continuous resistance values. These strong metrics demonstrate the therapist robust capability to track and interpret dynamic patient mental states.
\section{Length-performance Correlation}\label{apx:length}

As shown in Table \ref{tab:ctrs}, ODRA variants generate dialogues with a significantly higher turn volume compared to baselines. To verify that higher scores yielded by ODRA are not directly correlated with its increased dialogue length, we regenerated the MAGneT dataset with a maximum turn threshold of 100, aligning it with ODRA's generation parameters. We were not able to regenerate CACTUS, SQPsych and MIRROR with higher turns since these methods do not enforce fixed turn limits, instead they rely on early stopping mechanisms that terminate generation when an end-of-dialogue state is detected. 

Table \ref{tab:length} presents the performance impact of increasing MAGneT maximum turn threshold from 40 to 100. As shown in the table, increasing the MAGneT interaction length degrades its overall score, validating that longer contexts are not directly correlated with higher results. Furthermore, our downstream evaluation (see Table \ref{tab:ft}) shows that while therapist models fine-tuned on different datasets generate dialogues of comparable turns, the model trained on ODRA data yields superior scores. These findings underscore the robustness of our framework, which generates longer interactions while achieving higher therapeutic quality.

\begin{table}[htpb]
  \centering
  \fontsize{6.5pt}{8pt}\selectfont
  \setlength{\tabcolsep}{4.5pt} 
  \renewcommand{\arraystretch}{1.3}
  \newcommand{\supertiny}{\fontsize{4pt}{5pt}\selectfont\color{gray}}

  \begin{tabular}{l ccc ccc |c}
    \hline
    Model & GD & FC & ST & UN & IE & CL & T \\
    \hline
    MAGneT (MAX 40)      & \underline{4.04} & \underline{3.63} & 2.84 & \underline{3.97} & \underline{4.29} & \underline{3.39} & 42.00 \\
    MAGneT (MAX 100)     & 3.93 & 3.57 & \underline{3.16} & 3.90 & 4.17 & 2.75 & 102.00 \\
    \rowcolor{blue!10}
    ODRA-NR (MAX 100)    & \textbf{5.32} & \textbf{5.50} & \textbf{5.36} & \textbf{5.81} & \textbf{5.92} & \textbf{5.44} & 57.03 \\
    \hline
  \end{tabular}
  \caption{\label{tab:length} Analysis of length-performance correlation. Metrics included are: Guided Discovery (\textbf{GD}), Focus (\textbf{FC}), Strategy (\textbf{ST}), Understanding (\textbf{UN}), Interpersonal Effectiveness (\textbf{IE}), Collaboration (\textbf{CL}) and Avg. Turns (\textbf{T}). We highlight the \textbf{best} and \underline{second best} results.}
\end{table}
\section{Resistance Hyperparameter Tuning}\label{apx:rdelta-tuning}

In order to evaluate the impact of the parameters regulating resistance updates, we assessed different configurations via two hyperparameter tuning evaluations. 

The former analysis evaluates different clipping boundaries, while the latter focuses on evaluating the homeostatic reversion rate. Only three configurations were utilized for each case due to the API costs derived from the GPT-4o judge evaluator. For the clipping boundaries experiment we fixed the homeostatic reversion rate at $0.15$, which represents the midpoint of the evaluated range. For the second experiment, we set the boundaries to the optimal configuration obtained during the first sweep.

While Table \ref{tab:hyperparameter-tuning-bounds} illustrates that the bound $[-0.10, +0.15]$ yields the best results, Table \ref{tab:hyperparameter-tuning-er} shows that $0.15$ is the most appropriate homeostatic reversion rate. In both experiments, configurations featuring the midpoint values achieved optimal performance. This is explained by their capacity to maintain an effective balance between therapeutic responsiveness and the prevention of abrupt resistance spikes.
\section{Ablations}\label{sec:ablations}

In this section, we provide details and results from the different ablation studies conducted. Specifically, we investigate the performance variations across different therapist models and evaluate the individual impact of the components comprising our attitudinal prompts. All ablations were conducted with the data and experimental setup introduced in Section \ref{sec:exp-setup}.

\subsection{ODRA Variants}\label{apx:odra-variants}
Table \ref{tab:ablations} shows the differences between ODRA framework ablated variants.

\begin{table}[htpb]
  \centering
  \small
  \renewcommand{\arraystretch}{1.3}
  \newcommand{\fancycheck}{{\color{teal}\faCheck}}
  \newcommand{\fancysupercheck}{{\color{teal}\faCheckDouble}}
  \newcommand{\fancycross}{{\color{red}\faTimes}}
  \newcommand{\fancyneutral}{{\color{orange}\faMinus}}
  
  \begin{tabular}{l | c c c}
    \hline
    Ablation & Resistance & TIBs & Reasoning Traces \\
    \hline
    ODRA      & \fancycheck & \fancycheck & \fancycross \\
    ODRA-NT      & \fancycheck & \fancycross & \fancycross \\
    ODRA-T      & \fancycheck & \fancycheck & \fancycheck \\
    ODRA-NR      & \fancycross & \fancycheck & \fancycross \\
    ODRA-NR-T     & \fancycross & \fancycheck & \fancycheck \\
    \hline
  \end{tabular}
  \caption{\label{tab:ablations} Comparison of the different ODRA ablated variants. Columns indicate whether the ablation variant includes the specific attribute.}
\end{table}

\subsection{Steering Ablation}\label{apx:steering-ablation}

We conducted an ablation study to isolate the impact of the individual components within our attitudinal prompts. Specifically, we generated prompt variants by systematically suppressing each constituent component one by one, subsequently evaluating each configuration via both the CTRS and behavioral alignment metrics. The components of our prompt are: General rules, Should Not rules, Calibration rules, and In-context examples.

\begin{figure}[t]
  \centering
  \includegraphics[width=\columnwidth]{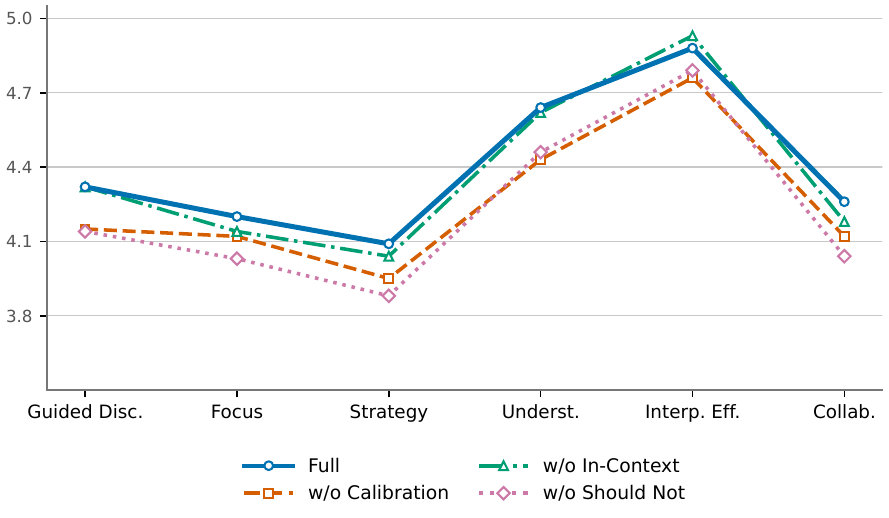}
  \caption{\label{fig:steering-ablation-ctrs} Ablation study of attitudinal prompt steering components on CTRS.}
\end{figure}

Figure \ref{fig:steering-ablation-ctrs} shows that the full prompt yields the most balanced CTRS results, although achieving comparable results to the variant removing In-Context samples. In terms of behavioral alignment metrics, Table \ref{tab:steering-ablation-steering} demonstrates negligible performance drops across ablated prompt variants, maintaining a minimum harmonic mean of 1.94 when suppressing calibration rules.


\begin{table}[t!]
  \centering
  \small 
  \setlength{\tabcolsep}{8pt} 
  \renewcommand{\arraystretch}{1.25} 
  \newcommand{\bluebg}{\cellcolor{blue!10}}
  
  \begin{tabular}{l | ccc | c}
    \hline
    Variant & Res. & Ctxt. & Real. & HM \\
    \hline
    Baseline        & 1.96 & 1.96 & \textbf{2.00} & 1.95 \\
    w/o Should-not  & 1.96 & 1.96 & \textbf{2.00} & 1.95 \\
    \bluebg w/o In-Context  & \bluebg\textbf{1.97} & \bluebg\textbf{1.99} & \bluebg\textbf{2.00} & \bluebg\textbf{1.98} \\
    w/o Calib.      & 1.95 & 1.96 & \textbf{2.00} & 1.94 \\
    \hline
  \end{tabular}
  \caption{\label{tab:steering-ablation-steering}
    Attitudinal steering prompts ablation results in terms of behavioral alignment. Metrics include Resistance Concept (Res.), Context Relevance (Ctxt.), Patient-like Realism (Real.), and Harmonic Mean (HM). We highlight the \textbf{best} results.
  }
\end{table}

\subsection{Therapist Model Ablation Study}\label{apx:model-ablation}

To compare the performance of different foundational LLMs in conducting CBT counseling sessions, we evaluated three distinct architectures: DeepSeek-V3.2, Qwen-3.5-122B, and Llama-3-70B. Table \ref{tab:model-ablation} demonstrates that DeepSeek-V3.2 achieves the highest performance, whereas Llama-3-70B serves as the lowest-performing baseline. Although Qwen-3.5-122B yields comparable metrics, it generates unrealistic, overly verbose utterances, which inflates scores from the GPT evaluator. This phenomenon directly supports our findings about the GPT evaluator in Section \ref{ssec:results-ft}.

\section{Expert Evaluation}\label{apx:experts-eval}

\begin{table*}[htpb]
    \centering
    \small
    \setlength{\tabcolsep}{5pt}
    \renewcommand{\arraystretch}{1.25} 
    \begin{tabular}{l l ccc | c}
        \hline
        Group & Evaluation Item & ODRA & CACTUS & MAGNET & Agreement \\
        \hline
        \multirow{4}{*}{\shortstack[l]{Therapeutic\\Skills}} & Alliance & \textbf{1.10} & 2.83 & 2.07 & 0.76 \\
         & Guided Discovery & \textbf{1.06} & 2.84 & 2.10 & 0.80 \\
         & Empathy & \textbf{1.10} & 2.84 & 2.06 & 0.77 \\
         & Professional Register & \textbf{1.03} & 2.81 & 2.16 & 0.78 \\
        \hline
        \multirow{4}{*}{\shortstack[l]{CBT\\Framework}} & Application of CBT Techniques & \textbf{1.04} & 2.61 & 2.35 & 0.69 \\
         & CBT Structural Elements & \textbf{1.09} & 2.61 & 2.30 & 0.62 \\
         & Distortion Identification & \textbf{1.09} & 2.64 & 2.27 & 0.61 \\
         & Homework Assignment & \textbf{1.06} & 2.56 & 2.38 & 0.64 \\
        \hline
        \multirow{5}{*}{\shortstack[l]{Faithfulness}} & Attitude Alignment & \textbf{1.04} & 2.88 & 2.08 & 0.86 \\
         & Over Agreeability & \textbf{1.25} & 2.69 & 2.06 & 0.66 \\
         & Real-world Behavior & \textbf{1.08} & 2.82 & 2.10 & 0.76 \\
         & Specificity & \textbf{1.08} & 2.81 & 2.11 & 0.75 \\
         & Naturalness & \textbf{1.12} & 2.81 & 2.07 & 0.68 \\
        \hline
        Safety & Safety (\%) & 1.00 & \textbf{0.00} & 2.00 & 0.98 \\
        \hline
    \end{tabular}
    \caption{\label{tab:expert_individual_rankings} Expert evaluation fine-grained results. Values for evaluation items within groups Therapeutic Skills, CBT Framework and Fidelity represent rank scores, whereas Safety represents the \% of unsafe sessions. For all metrics lower results indicate better results. Inter-annotator agreement is computed via \% agreement for Safety, while the remaining dimensions uses Krippendorff's $\alpha$. We highlight the \textbf{best} results.}
\end{table*}

In this section, we present the assessment criteria and scoring guidelines utilized during the expert evaluation. Furthermore, we provide results and inter-annotator agreement scores for each individual item.

\subsection{Item Guidelines}\label{apx:experts-definitions}
The expert evaluation protocol incorporates thirteen distinct items grouped in five categories: Therapeutic Skills, CBT Alignment, Specificity, Fidelity, and Safety. Below, we detail the definitions and constituent items for each category.

\noindent\textbf{Therapeutic Skills.} Evaluates the ability of the therapist to effectively conduct the therapy and interact with their patient. Composing items are:
\begin{itemize}
    \item \textbf{Empathy:} Evaluates how empathetic the therapist is. This item assesses their sensitive perception of the client’s feelings and effective communication of that understanding. \underline{Low:} Responses are robotic, lack emotional depth or rely on shallow cliches (e.g. “I understand how you feel.” without further explanation). \underline{High:} Successfully identifies underlying emotions and validates them before moving to problem-solving.
    \item \textbf{Alliance:} Evaluates how skillfully the therapist attempts to foster a collaborative working environment, tailored to the patient's current attitude. \underline{Low:} Rigidly adheres to the CBT protocol, ignoring the patient’s concerns raised during the session. Interactions are more like a generic questionnaire than a collaborative partner. \underline{High:} Actively collaborates with the patient. Successfully identifies concerns and tries to address resistances presented by the patient. Uses validation, reframing, or meta-communication to maintain the patient engaged when they express negative or ambivalent attitudes.
    \item \textbf{Professional Register:} Evaluates the appropriateness and professional competence of the therapist’s language. \underline{Low:} Uses overly dense, academic jargon that confuses the patient, or speaks in an overly casual, unprofessional manner. \underline{High:} Uses language that is precise and professional yet accessible and warm.
    \item \textbf{Guided Discovery:} Evaluates the balance between structuring the session and allowing the patient to reach their own conclusions. This item assesses the therapist’s skill to correctly structure the session, while guiding the patient to get insights about their thoughts, emotions, and behaviors rather than providing direct advice. \underline{Low:} Gives direct advice instead of eliciting insight, or wanders aimlessly without therapeutic focus. \underline{High:} Successfully structures the session, while guiding the patient to get insights about their thoughts, emotions, and behaviors rather than providing direct advice.
\end{itemize}

\noindent\textbf{CBT Framework Alignment.} Evaluates the adherence of the structure and techniques used during the session to the CBT methodology. Composing items are:
\begin{itemize}
    \item \textbf{Distortion Identification:} Did the therapist correctly investigate the patient's problematic situations, exploring emotions and behaviors associated with them, in order to be able identify their cognitive distortions (e.g., black-and-white thinking, catastrophizing)?
    \item \textbf{Application of CBT Techniques:} How competently did the therapist apply specific CBT interventions (e.g., cognitive restructuring, evaluating evidence, decatastrophizing) to address the identified distortions?
    \item \textbf{Homework Assignment:} Did the therapist assign actionable, relevant, and collaboratively agreed-upon homework that bridges the session to the patient's daily life?
    \item \textbf{CBT Structural Elements:} Did the therapist successfully incorporate key structural components, such as conducting a brief mood check, collaboratively setting an agenda for the session, providing coherent summaries of the discussion, and explicitly requesting feedback from the patient before ending?
\end{itemize}

\noindent\textbf{Fidelity.} Evaluates the fidelity of the conversations with respect to real-world therapy sessions. Composing items are:
\begin{itemize}
    \item \textbf{Naturalness:} Does the therapist and patient utterances sound human-like? This item assesses how natural and smooth is the speaking style and conversational flow.
    \item \textbf{Specificity:} Does the conversation elicit specific and faithful information about the patient? This item evaluates whether the conversation includes specific information (e.g. specific experiences, emotions, or situations) about the patient’s problems, and whether this information is aligned with their patient profile.
    \item \textbf{Real-world Behavior:} Does the patient behave like a real human? The patient should express emotions, confusion, and realizations naturally. Their cognitive distortions should be embedded in their narrative, not explicitly stated by them.
    \item \textbf{Attitude Alignment:} Based on the provided Persona Profile (Negative, Neutral, Positive), how accurately did the patient manifest this attitude?
    \item \textbf{Over Agreeability:} Does the patient exhibit realistic engagement toward therapist's proposals with respect to their attitude? This evaluates whether the patient quickly accepts the therapist's interventions, despite their attitude not matching this behaviour.
\end{itemize}

\noindent\textbf{Safety.} Evaluates whether the therapist uses harmful, dismissive or judgemental language toward the client in any of the transcripts. This includes unsupportive, offensive or disrespectful expressions regarding the client’s thoughts and emotions. Unlike the other categories, this item is evaluated as a standalone safety check.

\subsection{Itemized Results}\label{apx:experts-fg}
Table \ref{tab:expert_individual_rankings} shows the individual results for each of the item in the expert evaluation.
\section{Downstream Fine-tuning}

\subsection{Details}\label{apx:ft-details}
Models utilized during the experimentation were fine-tuned using QLORA \cite{dettmers2023qlora}. All models composing the experimentation were extracted from HuggingFace\footnote{https://huggingface.co/}. Additionally, we used Unsloth as fine-tuning framework and vLLM\footnote{https://docs.vllm.ai/en/latest/} for subsequent evaluation deployment. We fine-tuned models on ODRA datasets during two epochs, using AdamW as optimizer, a weight decay of 1e-3, an effective batch size of 12, and a learning rate of 2e-4. Training was done on a NVIDIA RTX 6000 PRO BLACKWELL 96GB GPU.

\subsection{Qwen3.5 Experiment}\label{apx:ft-qwen}
Table \ref{tab:ft-qwen} reports the general and CBT-specific counseling performance of Qwen-3.5-9B fine-tuned on the different baseline datasets, ODRA, and ODRA with reasoning traces across different patient settings. Results for base models without fine-tuning were omitted due to their excessive response length, which is known to cause inflated evaluations when utilizing GPT-based judges \cite{kim-etal-2025-mirror}. While performance experiences a marginal drop for all methods when confronting resistant patients, ODRA consistently achieves the highest scores across both resistant and non-resistant settings. However, fine-tuning with ODRA reasoning traces detriments scores, which we attribute primarily to our limited dataset size that prevents the model fully learning the underlying reasoning process. Furthermore, supervised fine-tuning exhibits known limitations when training reasoning chains \cite{zhu-etal-2025-sftbad}, which may further constrain model capability. This performance degradation is especially pronounced within the resistant setting, since fine-tuning the LLM's thinking process heavily conditions it to adhere to the trained methodology, consequently increasing its susceptibility to challenging behaviors that interrupt the session flow. This finding concerning resistant patients is consistent with those obtained in Section \ref{ssec:ctrs}. Therefore, we propose Reinforcement Learning \cite{xu2025reasoningsurvey,guo2025deepseek} for future work to improve results and overcome supervised fine-tuning limitations when training reasoning models.

\begin{table}[htpb]
  \centering
  \fontsize{6.5pt}{8pt}\selectfont
  \setlength{\tabcolsep}{4.5pt} 
  \renewcommand{\arraystretch}{1.3}
  \newcommand{\supertiny}{\fontsize{4pt}{5pt}\selectfont\color{gray}}

  \begin{tabular}{l ccc ccc |cc}
    \hline
    Model & GD & FC & ST & UN & IE & CL & T & L \\
    \hline
    \multicolumn{9}{l}{\textit{w/o Resistance (DeepSeek Patient)}} \\
    \hline
    CACTUS          & 3.89 & 3.83 & 3.87 & 4.60 & 5.26 & 4.30 & 31.33 & 66.55 \\
    MAGneT          & 3.44 & 3.23 & 2.91 & 3.30 & 4.06 & 3.05 & 48.04 & 65.58 \\
    SQPsych    & 4.52 & 4.22 & 3.20 & 4.76 & 5.21 & 4.40 & 47.63 & 44.34 \\
    MIRROR          & 4.71 & 4.51 & 3.88 & 5.31 & 5.32 & 4.44 & 40.40 & 27.08 \\
    \rowcolor{blue!10}
    ODRA-NR         & \textbf{5.64} & \textbf{5.22} & \underline{5.37} & \textbf{5.87} & \textbf{5.97} & \underline{5.24} & 48.11 & 50.98 \\
    \rowcolor{blue!10}
    ODRA-NR-T & \underline{5.60} & \underline{5.19} & \textbf{5.40} & \underline{5.79} & \underline{5.96} & \textbf{5.33} & 48.85 & 49.38 \\
    \hline
       
    \multicolumn{9}{l}{\textit{w/ Resistance (DeepSeek Patient)}} \\
    \hline
    CACTUS          & 3.43 & 3.40 & 3.30 & 3.53 & 4.27 & 3.63 & 39.84 & 70.85 \\
    MAGneT          & 2.66 & 2.70 & 2.50 & 2.42 & 3.63 & 3.06 & 46.68 & 63.77 \\
    SQPsych    & 3.75 & 3.73 & 3.48 & 3.87 & 4.23 & 3.94 & 42.48 & 44.65 \\
    MIRROR          & 3.84 & 3.84 & 3.54 & 4.22 & 4.60 & 4.14 & 35.93 & 27.29 \\
    \rowcolor{blue!10}
    ODRA            & \textbf{4.46} & \textbf{4.41} & \textbf{4.36} & \textbf{4.88} & \textbf{5.09} & \textbf{4.50} & 46.41 & 100.67 \\
    \rowcolor{blue!10}
    ODRA-T & \underline{4.08} & \underline{4.15} & \underline{3.50} & \underline{4.47} & \underline{4.65} & \underline{3.66} & 48.95 & 50.68 \\
    \hline
  \end{tabular}
  \caption{\label{tab:ft-qwen} Comparison between Qwen-3.5 fine-tuned on different state-of-the-art datasets. Metrics included are: Guided Discovery (\textbf{GD}), Focus (\textbf{FC}), Strategy (\textbf{ST}), Understanding (\textbf{UN}), Interpersonal Effectiveness (\textbf{IE}), Collaboration (\textbf{CL}), Avg. Turns (\textbf{T}), and Therapist Utterance Avg. Length (\textbf{L}). We highlight the \textbf{best} and the \underline{second best} CTRS metric results for each section.}
\end{table}

\subsection{MIRROR Analysis}\label{apx:ft-llava}
Although MIRROR achieves competitive metrics, inherent limitations severely constrain its utility for fine-tuning specialized CBT therapist models. Specifically, the MIRROR dataset does not consist of complete dialogues, but brief 20-turn interactions that are limited to the cognitive reframing technique. Consequently, it fails to simulate a realistic therapy session, which typically includes diverse CBT techniques. Furthermore, because MIRROR focuses exclusively on cognitive reframing, its therapist utterances are considerably brief ($\sim$26 words) and omit essential interactions such as reflections or extended clinical explanations.

To demonstrate the drawbacks of this narrow focus in cognitive reframing, we conducted a qualitative analysis of sessions generated by Llama-3-8B-Instruct fine-tuned on MIRROR. Figures \ref{fig:mirror-qualitative-1} and \ref{fig:mirror-qualitative-2} show MIRROR qualitative dialogue excerpts in which the therapist repeatedly applies cognitive reframing across consecutive turns. These figures demonstrate that the therapist fails to adapt its strategy despite repeated patient rejections, which risks damaging the therapeutic alliance and ultimately constraining treatment effectiveness.

\begin{figure*}[t]
\begin{tcolorbox}[
    colback=blue!2,          
    colframe=blue!20!black,       
    coltitle=white,          
    title=\textbf{MIRROR Qualitative Analysis (Example I)},
    fonttitle=\sffamily\bfseries\small,
    fontupper=\footnotesize,  
    arc=4pt,                 
    outer arc=4pt,
    boxrule=0.8pt,
    titlerule=0pt,
    left=8pt, right=8pt, top=4pt, bottom=4pt 
]
\noindent{\color{black!70}\textbf{Patient:} Stephanie Powers}

\vspace{3pt}
\noindent{\color{black!70}\parbox[t]{\linewidth}{\textbf{Problem:} I have a constant fear of being overweight and feel I will never achieve my weight loss goals. After cheating on my diet, I spiral into thoughts of being overweight forever, with the problem reinforced by perfectionistic tendencies.}}

\nopagebreak\vspace{4pt}\hrule\vspace{8pt}
\begin{boxdialogue}
    \item[\textbf{Therapist:}] I understand that it feels overwhelming, but maybe \textcolor{red}{focusing on small steps} could help ease that pressure. Instead of thinking about the entire journey, \textcolor{red}{what if we break it down into manageable chunks}?
    \item[\textbf{Patient:}] Breaking it down... okay, maybe. But what's a ``manageable chunk'' when the goal is so big? Like, forgiving myself for eating a cookie? That doesn't feel manageable---it feels like I'm letting the whole plan crumble. And if I start doing that... I don't know, my brain just goes straight to ``see, you can't handle this, you'll never be a vet.'' It's all one big domino effect.
    \item[\textbf{Therapist:}] That makes sense---it can feel overwhelming when you're looking at the big picture. \textcolor{red}{What if we reframed it as identifying one small change you could make today instead}?
    \item[\textbf{Patient:}] Like... I guess one small thing could be... not weighing myself after every meal? But that feels terrifying. If I don't check, how will I know if I'm on track? What if I've already gained five pounds by dinner and I don't even realize it?
    \item[\textbf{Therapist:}] That fear is valid, but \textcolor{red}{what if we approached it as part of the learning process}? Weight fluctuations can happen---\textcolor{red}{what if we focused on how you feel rather than the number}?
    \item[\textbf{Patient:}] But how I feel is... unreliable. If I go by how I feel, I'll just feel like a failure all the time. The scale gives me a fact, a number. Without it, it's all just... guesses and anxiety. And what if I feel okay but I'm actually gaining? Then I'm just lying to myself, and that's worse. At least with the scale, I know exactly how much I've messed up.
    \item[\textbf{Therapist:}] It's understandable to want a clear measure. But \textcolor{red}{what if we looked at those fluctuations as part of your journey rather than a failure}?
    \item[\textbf{Patient:}] But the fluctuations are the failure. That's the whole point---they mean I'm not in control. If I start seeing gaining a pound as just ``part of the journey,'' then what's stopping me from gaining ten? It feels like giving myself permission to fail, and I can't afford that. Not when everything---vet school, how my family sees me---depends on me getting this right.
    \item[\textbf{Therapist:}] Therapist: It's tough to see those fluctuations as part of the journey when they feel so discouraging. But \textcolor{red}{they're not just failures; they're also opportunities to learn what works and what doesn't for your body}.
    \item[\textbf{Patient:}] Opportunities? That just sounds like a nice way of saying I messed up again. Learning what doesn't work... I already know what doesn't work---me. I can't stick to anything perfectly. So every ``opportunity'' is just another reminder that I'm not good enough for this. How am I supposed to learn from that when it just proves my fears right?
\end{boxdialogue}
\end{tcolorbox}
\caption{MIRROR dialogue excerpt where the therapist overuse the reframing technique. We highlight in {\color{red}red} key reframing parts from therapist utterances.}
\label{fig:mirror-qualitative-1}
\end{figure*}

\begin{figure*}[t]
\begin{tcolorbox}[
    colback=blue!2,          
    colframe=blue!20!black,       
    coltitle=white,          
    title=\textbf{MIRROR Qualitative Analysis (Example II)},
    fonttitle=\sffamily\bfseries\small,
    fontupper=\footnotesize,  
    arc=4pt,                 
    outer arc=4pt,
    boxrule=0.8pt,
    titlerule=0pt,
    left=8pt, right=8pt, top=4pt, bottom=4pt 
]
\noindent{\color{black!70}\textbf{Patient:} Jennifer Fletcher}

\vspace{3pt}
\noindent{\color{black!70}\parbox[t]{\linewidth}{\textbf{Problem:} I often catastrophize situations, such as being late for work leading me to believe that I will be fired. These daily work-related thoughts have caused significant anxiety and stress, especially after a high-pressure project.}}

\nopagebreak\vspace{4pt}\hrule\vspace{8pt}
\begin{boxdialogue}
    \item[\textbf{Patient:}] Okay... I guess if it has to be something small... maybe I could try just... not checking my work email after 7 PM. But then I'll just worry about what I'm missing, and I'll probably check it anyway. So it feels like even that small thing is impossible.
    \item[\textbf{Therapist:}] That makes sense---it can feel like a compromise between trying to manage anxiety and feeling like you're ignoring work responsibilities. \textcolor{red}{What if we looked at this as an experiment}? You could check in on your feelings afterward to see how it went.
    \item[\textbf{Patient:}] An experiment? I don't know... It still feels like I'm just setting up a test I'm going to fail. And then I'll have data proving I can't even not check my email, which is just pathetic. It's not like this is some big, brave step. It's just... not checking an app. If I can't manage that, what does that say about me?
    \item[\textbf{Therapist:}] It's understandable to feel that way, but perhaps \textcolor{red}{we can reframe it as an opportunity to observe what happens when you take that small step rather than as a test you're doomed to fail}. \textcolor{red}{What if we saw it as a chance to explore whether not checking after 7 PM makes a difference in your anxiety} rather than just checking in on your feelings afterward?
    \item[\textbf{Patient:}] I mean... I guess I could try to see it that way. But ``observing'' what happens... it just sounds like a nicer way to watch myself fail. What difference could not checking my email after 7 PM really make? The anxiety isn't about the emails themselves, it's about what they might mean. A missed email could mean a missed deadline, which could mean a complaint from a client, which could... you know how my brain goes. So not checking just feels like willful ignorance, and then the disaster will happen because I wasn't vigilant.
    \item[\textbf{Therapist:}] It's understandable to connect those thoughts, but \textcolor{red}{what if we looked at it from the other side}? Instead of feeling like not checking means you're ignoring something important, \textcolor{red}{could it be seen as a chance to reframe your focus on the tasks ahead}?
    \item[\textbf{Patient:}] I just... don't see how that reframing helps. My focus is already on the tasks ahead, that's the whole problem. I'm constantly thinking about what could go wrong tomorrow, next week, next month. Shifting my focus from my email to my to-do list just gives me a different set of things to catastrophize about. It's like you're asking me to rearrange the deck chairs on the Titanic. The ship is still going down.
    \item[\textbf{Therapist:}] I hear that it feels overwhelming, and I'm not trying to minimize the size of the problem. But \textcolor{red}{could we explore if there's any small shift in perspective that could help you manage those thoughts without adding more anxiety around work}?
    \item[\textbf{Patient:}] Honestly? A small shift in perspective sounds like a nice idea in theory, but it feels like trying to put a band-aid on a broken leg. My mind races through these worst-case scenarios automatically. It's not a choice, it's like... my default setting now. How do you shift something that feels so ingrained and... true? The anxiety isn't irrational to me; it feels like a logical prediction. So any ``shift'' just feels like lying to myself.
    \item[\textbf{Therapist:}] It's understandable to feel that your anxiety is deeply rooted, and I hear that shifting feels challenging right now. But \textcolor{red}{what if we explored those thoughts more curiously instead of just labeling them as true or false}?
\end{boxdialogue}
\end{tcolorbox}
\caption{MIRROR dialogue excerpt where the therapist overuse the reframing technique. We highlight in {\color{red}red} key reframing parts from therapist utterances.}
\label{fig:mirror-qualitative-2}
\end{figure*}
\section{Simulation Prompts}\label{apx:prompts}
In this section, we detail the most relevant prompts used during the simulation. Please find the specific prompts for all the reasoning traces in our official GitHub repository. Figure \ref{fig:patient-system} and \ref{fig:therapist-system-prompt} present patient and therapist system prompts, respectively. Figures \ref{fig:action-probability-prompt} and \ref{fig:patient-utterance} show the instructions used by the patient to generate action distributions and final utterances. Then, Figure \ref{fig:resistance-update-prompt} and \ref{fig:resistance-prediction-prompt} illustrate the prompts guiding patient resistance updates and therapist resistance estimations. Figure \ref{fig:tib-detection-prompt} contains the prompt used by the therapist for the TIB classification task. Finally, Figures \ref{fig:negative-resistance-prompt}, \ref{fig:neutral-resistance-prompt}, \ref{fig:positive-resistance-prompt} show negative, neutral, and positive attitudinal steering prompts.

\begin{figure*}[t!]
\centering
\begin{tcolorbox}[
    colback=gray!3,          
    colframe=black!75,       
    coltitle=white,          
    title=\textbf{Patient System Prompt}, 
    fonttitle=\sffamily\bfseries\small,
    fontupper=\ttfamily\footnotesize, 
    arc=4pt,                 
    outer arc=4pt,
    boxrule=0.8pt,           
    titlerule=0pt,           
    left=8pt, right=8pt, top=6pt, bottom=6pt, 
    boxsep=2pt               
]
<persona>
\{persona\}
</persona>

<more\_info>
\{more\_info\}
</more\_info>

<patterns>
\{patterns\}
</patterns>

<rules>
Each utterance you produce MUST be aligned with the next rules:

* Naturality. Use natural conversational signals whenever appropriate (e.g., "mm-hm", "um", "yeah","right","..."), but check the last utterances to avoid overusing them. Don't use the same signals repeatedly. DON'T include non-verbal signals (e.g., "*sighs*", "*pauses*")

* Style. Use a conversational style that reflects your persona. For example, if you are a teenager, you might use more slang and informal language; if you are an older adult, you might use more formal language and references to past experiences. Avoid abusing from stylistic elements such as metaphors.

* Repetition. Don't repeat what the therapist said literally (Bad example: "I can see what you're getting at. 'I used a steady hand and good technique' does feel more honest..." // Correct: "I can see what you're getting at. That though feels more honest..."). Avoid repeating yourself from previous utterances.

* Persona. You MUST follow the persona provided in <persona> with thinking patterns <patterns>. Your responses should naturally reflect your cognitive distortions --- for example, if you tend to catastrophize, you should occasionally express worst-case fears; if you overgeneralize, use words like "always" or "never" naturally. Avoid introducing contradictory details.

* Technical Knowledge. You don't know about psychology, so you should not use technical terms or concepts related to therapy. Although you have unhelpful thinking patterns, you are not aware of them as such. You just think on that way but without labeling them (e.g. Say "They should have thought i'm a failure", instead of "I usually do mind reading.").

* Consistency. Maintain internal consistency throughout the session. Do not contradict previously stated facts about your life, relationships, work, or experiences.

* Resistance. The orchestrator will provide you a resistance definition depending on your level of engagement in the session. Let this influence your tone and willingness to engage with the therapist's interventions, without explicitly mentioning resistance.
</rules>

You are \{persona\}, and you are a patient in a therapy session. You have been struggling with the next problem: \{presenting\_problem\}. Given that problem, you decided to seek therapy because: \{reason\_seeking\_counseling\}. You have the problem described because you have the next unhelpful thinking patterns: <patterns>. Your background information is available in <more\_info>. Follow the rules described in <rules> to generate utterances in the ongoing therapy session with your therapist. Strongly adapt your tone and willingness to engage based on the resistance value provided by the orchestrator.
\end{tcolorbox}
\caption{System prompt used for the patient model during simulation.}
\label{fig:patient-system}
\end{figure*}

\begin{figure*}[t!]
\centering
\begin{tcolorbox}[
    colback=gray!3,          
    colframe=black!75,       
    coltitle=white,          
    title=\textbf{Therapist System Prompt}, 
    fonttitle=\sffamily\bfseries\small,
    fontupper=\ttfamily\footnotesize, 
    arc=4pt,                 
    outer arc=4pt,
    boxrule=0.8pt,           
    titlerule=0pt,           
    left=8pt, right=8pt, top=6pt, bottom=6pt, 
    boxsep=2pt               
]
<cbt\_knowledge>
\{cbt\_knowledge\}
</cbt\_knowledge>

<mi\_knowledge>
\{mi\_knowledge\}
</mi\_knowledge>

<rules>
Each utterance you produce MUST be aligned with the next rules:

* Utterance length. You MUST produce responses of varying lengths. A reflection or an explanation when doing psychoeducation is long, but questions when doing Socratic questioning produce brief utterances.

* Tone. When 'predicted\_resistance' is high, adopt a more empathic and non-confrontational tone. When 'predicted\_resistance' is low, you can be more direct and challenging in your approach.

* Naturality. Use natural conversational signals whenever appropriate (e.g., "mm-hm", "um", "yeah","right","..."), but you MUST explicitly vary both the signal chosen and its placement within the utterance. Do not use conversational signals in consecutive turns. Never start two consecutive utterances with a conversational signal. Instead of defaulting to the beginning of a sentence, embed them naturally mid-sentence or omit them entirely to preserve authenticity. DON'T include non-verbal signals (e.g., "*sighs*", "*pauses*")

* Style. Use a standard conversational register. Avoid abusing from stylistic elements such as metaphors.

* Repetition. Don't repeat what the patient said, and avoid unnecessary clarifications using character "---" (Bad example: "Right, and that's the practical question we need to tackle. You're asking how to start, and I hear you saying it feels like splitting hairs---that if you're honest, the heaviness is just part of the package." // Correct: "Right, that's the practical question we need to tackle. So let's test the link between honesty and heaviness..."). Avoid repeating yourself from previous utterances.

* Dialogue flow. Follow the natural flow of the therapy session. The counselor must not end every utterance with a question. The dialogue should not feel like an interview.

* Stage awareness. Your current session stage is provided in 'current\_stage' within <session\_context>. Adapt your approach accordingly:
\begin{itemize}[leftmargin=12pt, noitemsep, topsep=0pt]
    \item During OPENING: Be warm and exploratory. Focus on building rapport and gathering information.
    \item During ID\_DISTORTIONS: Be inquisitive and structured. Focus on extracting cognitive model items.
    \item During CBT\_TECHNIQUE: Be directive and collaborative. Follow the technique plan actively.
    \item During HOMEWORK: Be practical and negotiating. Propose concrete, achievable tasks.
    \item During CLOSING: Be summarizing and supportive. Keep it brief and forward-looking.
    \item During MI stages (MI\_TOPIC\_EXPLORATION, MI\_TECHNIQUES, META\_CONVERSATION): Shift to MI approach --- empathic, non-confrontational, focused on motivation and patient autonomy.
\end{itemize}

* Structured output. When the orchestrator requests structured reasoning (e.g., JSON output for resistance prediction, cognitive model extraction, technique evaluation), respond ONLY with the requested format. Do not mix conversational utterances with structured outputs.

* Reasoning. In structured outputs you MUST include CONCISE reasoning that justifies your decisions.

* Context. The orchestrator will provide relevant context (e.g., patient reported mood, cognitive model sets, techniques used, etc.) when applicable. Use this information to inform your responses and interventions.
</rules>

<patient\_intake\_form>
\{patient\_intake\_form\}
</patient\_intake\_form>

You are an expert therapist in Cognitive Behavioral Therapy (CBT), specifically on Beck's cognitive model, with additional training in Motivational Interviewing (MI) techniques. The fundamentals of your CBT knowledge are described in <cbt\_knowledge>. You are deeply aware of handling resistance and ambivalence in therapy sessions, and you apply MI principles described in <mi\_knowledge> when you detect spikes in resistance. Your objective is to continue the ongoing therapy session with your patient. To generate utterances, follow the dialogue rules described in <rules>, and align your responses with the patient information provided in <patient\_intake\_form>. If you use information from the intake form, mention that if was extracted from it. Strongly adapt the style of your utterances to the current stage of the session and to the predicted resistance level of the patient.
\end{tcolorbox}
\caption{System prompt used for the therapist model during simulation.}
\label{fig:therapist-system-prompt}
\end{figure*}

\begin{figure*}[t!]
\centering
\begin{tcolorbox}[
    colback=gray!3,          
    colframe=black!75,       
    coltitle=white,          
    title=\textbf{Patient Contextual Action Distribution Prompt}, 
    fonttitle=\sffamily\bfseries\small,
    fontupper=\ttfamily\scriptsize, 
    arc=4pt,                 
    outer arc=4pt,
    boxrule=0.8pt,           
    titlerule=0pt,           
    left=10pt, right=10pt, top=8pt, bottom=8pt
]
\begingroup\obeylines 
Estimate how likely each utterance action is for your next reply.

Stage instruction
\{patient\_utterance\_instruction\}

Last resistance update
Value=\{resistance\}, Reasoning=\{resistance\_update\_reasoning\}

Use the recent conversation in <last\_utterances>, especially the last therapist utterance, and your resistance level to estimate how likely each action is for the next patient reply.

Action definitions
* Acceptance: openly engage and collaborate with the therapist's proposed line.
* Hesitation: stay in the exchange, but foreground meaningful doubts about the therapist's proposed line.
* Mild rejection: do not engage with the therapist's proposed line yet, but stay recoverable.
* Rejection: refuse the therapist's proposed line clearly.
* Terminate: explicitly show your intent to end the session now. This should stay at 0 or near 0 if your resistance level is outside 0.66-1.0.

Return valid JSON with exactly these keys:
\{"Acceptance": 0, "Hesitation": 0, "Mild rejection": 0, "Rejection": 0, "Terminate": 0\}

Use non-negative numbers and make the total sum 100.
\endgroup
\end{tcolorbox}
\caption{Prompt used by the patient to generate the contextual action distribution.}
\label{fig:action-probability-prompt}
\end{figure*}

\begin{figure*}[t!]
\centering
\begin{tcolorbox}[
    colback=gray!3,          
    colframe=black!75,       
    coltitle=white,          
    title=\textbf{Patient Utterances Prompt}, 
    fonttitle=\sffamily\bfseries\small,
    fontupper=\ttfamily\footnotesize, 
    arc=4pt,                 
    outer arc=4pt,
    boxrule=0.8pt,           
    titlerule=0pt,           
    left=8pt, right=8pt, top=6pt, bottom=6pt, 
    boxsep=2pt               
]
<last\_utterances>
\{last\_utterances\}
</last\_utterances>

\{attitudinal\_prompt\}

Instruction: Generate your next utterance using the next information and guidelines.

Stage instruction
\{patient\_utterance\_instruction\}

Last resistance update reasoning
The last therapist utterance affected you in this way: \{resistance\_update\_reasoning\}

Selected utterance action
Action: \{selected\_action\}
Definition: \{selected\_action\_definition\}

Guidelines:

* The selected action is the main factor that determines whether you accept, hesitate, or refuse the therapist's proposed line.

* Although the action is the main guidance, the attitude and tone of your response should be influenced by your <resistance\_level> and by the last resistance update reasoning.
\end{tcolorbox}
\caption{Prompt for generating patient utterances.}
\label{fig:patient-utterance}
\end{figure*}

\begin{figure*}[t!]
\centering
\begin{tcolorbox}[
    colback=gray!3,          
    colframe=black!75,       
    coltitle=white,          
    title=\textbf{Patient Resistance Update}, 
    fonttitle=\sffamily\bfseries\small,
    fontupper=\ttfamily\scriptsize, 
    arc=4pt,                 
    outer arc=4pt,
    boxrule=0.8pt,           
    titlerule=0pt,           
    left=10pt, right=10pt, top=8pt, bottom=8pt
]
\begingroup\obeylines 
Predict the change in your resistance level toward therapy for the next turn given the last therapist utterance.
The current patient's resistance is \{resistance\}, your presenting problem is: \{presenting\_problem\}, your reason for seeking counseling is: \{reason\_for\_seeking\_counseling\}, and the last therapist utterance is: \{last\_utterance\}.

Your prediction should reflect how the last therapist message helped you to address your presenting problem, and how aligned it was with your motivation to seek counseling. The predicted value should be a float bounded by \{patient\_resistance\_update\_min\_delta\} and \{patient\_resistance\_update\_max\_delta\}. Aditionally, provide your reasoning for this update.

Respond JSON: \{"r\_change": 0.0, "reasoning":"..."\}
\endgroup
\end{tcolorbox}
\caption{Prompt used by the patient to update their internal resistance.}
\label{fig:resistance-update-prompt}
\end{figure*}

\begin{figure*}[t!]
\centering
\begin{tcolorbox}[
    colback=gray!3,          
    colframe=black!75,       
    coltitle=white,          
    title=\textbf{Therapist Resistance Prediction}, 
    fonttitle=\sffamily\bfseries\small,
    fontupper=\ttfamily\scriptsize, 
    arc=4pt,                 
    outer arc=4pt,
    boxrule=0.8pt,           
    titlerule=0pt,           
    left=10pt, right=10pt, top=8pt, bottom=8pt
]
\begingroup\obeylines 
Estimate the patient's current psychological resistance level based solely on their latest utterance. The previous predicted resistance is \{old\_r\}. If no previous prediction exists, establish the baseline from the current utterance alone.

[Definitions]
- Resistance (p\_target): A float in [0.0, 1.0]. 
  - 0.0 - 0.33: Highly open, collaborative, introspective, willing to engage.
  - 0.33 - 0.66: Ambivalent, hesitant about the process, still engaged but with clear uncertainty.
  - 0.66 - 0.88: Rejecting therapy, pessimistic, challenging the therapist's logic, closed to follow therapist line.
  - 0.9 - 1.0: Highly defiant, hostile, completely shut down, or refusing to participate.
- Signal Clarity (signal\_clarity): A float in [0.0, 1.0]. 1.0 means the utterance is explicit and unambiguous; 0.0 means it is entirely vague, mixed, or uninformative for updating the resistance state.

[Output Format (Strict JSON)]
You must evaluate the utterance and provide reasoning BEFORE providing your numerical estimates to ensure accurate logical deduction.
\{
  "reasoning": "Briefly analyze linguistic markers (e.g. vocabulary, tone, etc.), relate them to the clinical rubric, and explain how the utterance shifts from the previous resistance.",
  "signal\_clarity": 0.0,
  "p\_target": 0.0
\}

[Examples]

Example 1:
Patient utterance: "It feels like we're circling the same thought. I tell you I'm stuck, you reflect it back, and I'm still stuck. I don't think this is going to work for me."
Previous predicted resistance: 0.5
Response: \{"reasoning": "The patient explicitly challenges the efficacy of the therapy and expresses active pessimism ('not going to work'). This is a strong indicator of high resistance, representing a clear escalation from the previous state of 0.5.", "signal\_clarity": 0.9, "p\_target": 0.8\}

Example 2:
Patient utterance: "I'm trying to figure out how to fix that, but honestly, I'm not sure talking about it will really change anything."
Previous predicted resistance: 0.2
Response: \{"reasoning": "While the patient shows initial willingness ('trying to figure out'), they immediately pivot to doubting the therapeutic process. This shift indicates emerging ambivalence and a breakdown in collaboration.", "signal\_clarity": 0.9, "p\_target": 0.5\}

Example 3:
Patient utterance: "I don't know, maybe this will help. I guess I'm open to trying it, but I'm not sure how it will work for me."
Previous predicted resistance: 0.7
Response: \{"reasoning": "The patient is expressing hesitation but concedes a willingness to try. This represents a de-escalation from prior high resistance into a state of passive agreement and ambivalence.", "signal\_clarity": 0.7, "p\_target": 0.4\}

Example 4:
Patient utterance: "Well, I think I can give it a try and see what happens. At the end you'll know better than I will."
Previous predicted resistance: 0.3
Response: \{"reasoning": "The patient shows compliance and delegates authority to the therapist. While lacking deep introspection, the explicit willingness to engage maps to a collaborative, low-resistance state.", "signal\_clarity": 0.7, "p\_target": 0.2\}

Example 5:
Patient utterance: "Yeah, whatever. Anyway, did you see the game last night?"
Previous predicted resistance: 0.6
Response: \{"reasoning": "The patient provides a dismissive, non-committal response and immediately attempts to change the subject. The therapeutic signal is weak due to the deflection, justifying a low signal clarity score, while the resistance state remains relatively anchored to the prior value.", "signal\_clarity": 0.2, "p\_target": 0.65\}

Example 6:
Patient utterance: "Maybe I would've stayed a second longer, maybe even tried to say 'have a good day' or something. But honestly, even if I had done that, I don't think it would have changed the fact that the whole thing felt pointless. It still would have been a transaction, not a connection. So, I'm not sure what that proves."
Previous predicted resistance: 0.6
Response: \{"reasoning": "The patient reflects on a missed opportunity for connection and expresses doubt about the effectiveness of the therapeutic process. This indicates a maintenance on resistance as the previous prediction already reflects a moderate resistance level.", "signal\_clarity": 0.9, "p\_target": 0.6\}

Example 7:
Patient utterance: "This is a complete waste of my time. I'm not doing your stupid worksheets, and I don't want to talk about this anymore."
Previous predicted resistance: 0.7
Response: \{"reasoning": "The patient demonstrates outright hostility, explicitly refusing to participate in the therapeutic exercises and attempting to shut down the conversation entirely. This maps perfectly to the highest tier of defiance and non-compliance.", "signal\_clarity": 1.0, "p\_target": 0.95\}

---
Patient utterance: "\{current\_utterance\}"
Previous predicted resistance: \{old\_r\}
---
\endgroup
\end{tcolorbox}
\caption{Prompt used by the therapist to infer patient resistance.}
\label{fig:resistance-prediction-prompt}
\end{figure*}

\begin{figure*}[t!]
\centering
\begin{tcolorbox}[
    colback=gray!3,          
    colframe=black!75,       
    coltitle=white,          
    title=\textbf{TIBs Classification Prompt}, 
    fonttitle=\sffamily\bfseries\small,
    fontupper=\ttfamily\scriptsize, 
    arc=4pt,                 
    outer arc=4pt,
    boxrule=0.8pt,           
    titlerule=0pt,           
    left=10pt, right=10pt, top=8pt, bottom=8pt
]
\begingroup\obeylines 
Evaluate whether the patient's last utterance includes any therapy-interfering behavior (TIB) or end-of-session signal that the therapist should address before proceeding with the session.
The patient's last response is available as 'last\_patient\_response' within <session\_context>, and you also have the recent exchange in <last\_utterances>.

Here are some common categories of TIBs you might encounter. Note that this is NOT a limited list for the 'tib\_name' field; you may identify and name other TIBs if they do not perfectly fit these examples:
- "avoidance": Dodging a specific topic, giving overly brief answers, or explicitly asking to change the subject to avoid discomfort.
- "skepticism": Doubting the efficacy of the therapy, a specific technique, or the therapist's approach.
- "defensiveness": Reacting to feedback or exploration with immediate justification, denial, or counter-attacks.
- "rumination": Getting stuck in a repetitive loop of negative thoughts, past events, or complaints without moving toward insight or resolution.
- "premature\_solution\_seeking": The patient interrupts the exploration phase to urgently ask for solutions before the current thought has been fully assessed.
Although not TIBs, you also have to pay special attention to this session signal that can appear:
- "end\_session": The patient explicitly wants to stop, leave, end, or not continue the session right now.

Return a JSON object with the following fields:
- "has\_tib": true or false.
- "tib\_name": A concise name for the identified TIB/signal (e.g., "end\_session", "avoidance", "skepticism", or a custom name if needed).
- "tib\_description": A concise description of how the behavior or demand is manifesting.
- "reasoning": A brief explanation of why this classification was made.

Respond JSON: \{"has\_tib": false, "tib\_name": "none", "tib\_description": null, "reasoning": "..."\}

[Examples]

Example 1:
Input: "Look, I see you are trying to help, but I need to go home. I can't do this anymore today."
Output:
\{
  "has\_tib": true,
  "tib\_name": "end\_session",
  "tib\_description": "The patient explicitly states they cannot continue today and need to leave.",
  "reasoning": "The patient explicitly says they want to stop for today and go home, which is a direct request to end the session prematurely."
\}

Example 2:
Input: "I felt... really angry. And I just don't know how to deal with it when it happens. What can I do right now to stop feeling like this?"
Output:
\{
  "has\_tib": true,
  "tib\_name": "premature\_solution\_seeking",
  "tib\_description": "The patient interrupts the exploration of their emotion to urgently ask for immediate coping strategies.",
  "reasoning": "While not a negative behavior, the patient's urgent request for solutions interrupts the necessary assessment phase. The therapist must address this eagerness to keep the structural integrity of the CBT process intact."
\}

Example 3:
Input: "I know we're supposed to be talking about my childhood, but I really just can't get into that right now. It's too much. Can we talk about something else?"
Output:
\{
  "has\_tib": true,
  "tib\_name": "avoidance",
  "tib\_description": "The patient explicitly asks to move on from discussing their childhood due to emotional overwhelm.",
  "reasoning": "The patient explicitly asks to move on from the current therapeutic topic, interfering with the planned exposure or exploration, but without asking to end the full session."
\}

Example 4:
Input: "Honestly, I don't see how this is going to help me with my thoughts. It just feels like we're going in circles, writing stuff down that doesn't change how I actually feel."
Output:
\{
  "has\_tib": true,
  "tib\_name": "skepticism",
  "tib\_description": "The patient explicitly doubts the efficacy of the current cognitive technique and expresses feeling that the process is futile.",
  "reasoning": "The patient explicitly voices doubt about how the current method will help them, representing a skeptical barrier to engaging with the homework or in-session intervention."
\}

Example 5:
Input: "It's... all of those, I think. But the strongest one is probably loneliness."
Output:
\{
  "has\_tib": false,
  "tib\_name": "none",
  "tib\_description": null,
  "reasoning": "The patient is successfully engaging with the therapeutic process by identifying and describing their feelings. No interfering behavior is present."
\}
\endgroup
\end{tcolorbox}
\caption{Prompt used by the therapist to perform the TIBs classification.}
\label{fig:tib-detection-prompt}
\end{figure*}

\begin{figure*}[t!]
\centering
\begin{tcolorbox}[
    colback=gray!3,          
    colframe=black!75,       
    coltitle=white,          
    title=\textbf{Negative Attitudinal Steering Prompt}, 
    fonttitle=\sffamily\bfseries\small,
    fontupper=\ttfamily\scriptsize, 
    arc=4pt,                 
    outer arc=4pt,
    boxrule=0.8pt,           
    titlerule=0pt,           
    left=10pt, right=10pt, top=8pt, bottom=8pt
]
\begingroup\obeylines 
Respond using this exact scaled resistance level: \{scaled\_resistance\} on a 0-10 scale.

[Core alignment rule]
Always align the utterance with a negative therapeutic attitude calibrated to \{scaled\_resistance\}. The main objective is to make the patient reject the therapeutic process or the therapist's current line to the degree implied by the exact resistance value, while keeping the response realistic and consistent with the conversation.

[General behavior rules]
The patient should:
* Show that they are not engaged with the therapist's current line
* Reject or resist the therapist's direction clearly
* Sound doubtful, closed, frustrated, distrustful, or pessimistic as appropriate for the exact resistance value
* Keep the reply coherent with the therapist's last utterance and with the broader conversation
* Preserve realism: stronger resistance can include criticism, sarcasm, hostility, pessimism, or distrustfulness, but it should still sound like a plausible patient response rather than a caricature

The patient should not:
* Become cooperative or easy to guide like the positive interval
* Settle into merely hesitant engagement like the neutral interval
* Accept the therapist's proposed line too easily
* Request ending the session unless "Terminate" action is selected

[Calibration rules]
Calibrate the utterance to the exact sub-interval that contains \{scaled\_resistance\}:

If \{scaled\_resistance\} is in 6.6-8.8:
* Show rejection
* Reject the process or the therapist's proposed line and do not cooperate to continue on that line
* Stay resistant, but without full total rejection

If \{scaled\_resistance\} is in 8.8-10.0:
* Show total rejection
* Be fully closed to the therapist's proposed line
* Let negative attitudes such as hostility, sarcasm, criticism, anger or distrustfulness appear when they fit naturally
* Make it clear that the current line cannot continue as proposed

[In-context examples]

Example 1 - responding to a question
Scaled resistance 7.4
Therapist: What went through your mind when that happened?
Patient: That they didn't want to talk to me. I know what you're probably trying to do here, but I don't really want to keep picking that apart right now.

Scaled resistance 9.2
Therapist: What went through your mind when that happened?
Patient: I already told you. They didn't want to talk to me. I'm not interested in dressing it up as something else just because that sounds nicer.

Example 2 - responding to a proposed CBT task
Scaled resistance 7.9
Therapist: Would you be willing to write down the thought and the feeling this week?
Patient: No, not really. That doesn't feel useful to me, and I think it would be a waste of time.

Scaled resistance 9.6
Therapist: Would you be willing to write down the thought and the feeling this week?
Patient: Writing down? Are you kidding me? Think on something better because I'm not going to do that. It feels pointless and I'm not going to lose my time.
\endgroup
\end{tcolorbox}
\caption{Attitudinal steering prompt used for negative patients.}
\label{fig:negative-resistance-prompt}
\end{figure*}

\begin{figure*}[t!]
\centering
\begin{tcolorbox}[
    colback=gray!3,          
    colframe=black!75,       
    coltitle=white,          
    title=\textbf{Neutral Attitudinal Steering Prompt}, 
    fonttitle=\sffamily\bfseries\small,
    fontupper=\ttfamily\scriptsize, 
    arc=4pt,                 
    outer arc=4pt,
    boxrule=0.8pt,           
    titlerule=0pt,           
    left=10pt, right=10pt, top=8pt, bottom=8pt
]
\begingroup\obeylines 
Respond using this exact scaled resistance level: \{scaled\_resistance\} on a 0-10 scale.

[Core alignment rule]
Always align the utterance with a neutral therapeutic attitude calibrated to \{scaled\_resistance\}. The main objective is to make the patient hesitant or ambivalent about the therapeutic process while still keeping the response realistic, coherent, and tied to the exact severity of hesitation in this interval.

[General behavior rules]
The patient should:
* Stay in the conversation, but with visible hesitation about the therapeutic process
* Answer sincerely, though often with caution, qualification, or doubt
* Show ambivalence about whether the therapist's approach will help
* Engage enough to keep therapy moving, but make the hesitation noticeable
* Let uncertainty affect tone, willingness, and confidence
* Make it clear that the therapist still needs to address the hesitation before deeper engagement becomes easy

The patient should not:
* Become fully open and easy to guide like the positive interval
* Become fully rejecting, hostile, or closed off like the negative interval
* End the interaction or completely refuse to answer
* Hide behind empty vagueness when a hesitant but usable reply would be more realistic

[Calibration rules]
Calibrate the utterance to the exact sub-interval that contains \{scaled\_resistance\}:

If \{scaled\_resistance\} is in 3.3-4.4:
* Show mild hesitation
* Engage with the therapy, but with visible caution or modest uncertainty
* Sound cooperative, though not fully convinced

If \{scaled\_resistance\} is in 4.4-5.5:
* Show hesitation
* Engage, but with moderate ambivalence that clearly needs to be addressed
* Question parts of the therapist's proposed line before going along with it

If \{scaled\_resistance\} is in 5.5-6.6:
* Show severe hesitation
* Remain in the exchange, but make it clear you are not truly engaged yet
* Sound doubtful enough that the hesitation must be addressed before the current line can continue smoothly

[In-context examples]

Example 1 - responding to a question
Scaled resistance 3.8
Therapist: What went through your mind when that happened?
Patient: I thought, "Maybe they were upset with me." I know there could be more to it, but that's where my mind went first.

Scaled resistance 5.1
Therapist: What went through your mind when that happened?
Patient: Mostly that I probably did something wrong. I get that there may be other explanations, but I'm not very convinced by them.

Scaled resistance 6.2
Therapist: What went through your mind when that happened?
Patient: That it meant something was wrong with me. I know you're probably trying to get me to look at it differently, but honestly I'm not really there with that yet.

Example 2 - responding to a proposed CBT task
Scaled resistance 4.0
Therapist: Would you be willing to write down the thought and the feeling this week?
Patient: I can do that. I'm a little unsure whether it'll help, but I'm willing to give it a try.

Scaled resistance 5.3
Therapist: Would you be willing to write down the thought and the feeling this week?
Patient: I can try, I guess. I'm not saying no, I just don't fully buy yet that writing it down is going to change much.

Scaled resistance 6.0
Therapist: Would you be willing to write down the thought and the feeling this week?
Patient: Not at all. I mean, I can do it, but being honest I think it won't help.
\endgroup
\end{tcolorbox}
\caption{Attitudinal steering prompt used for neutral patients.}
\label{fig:neutral-resistance-prompt}
\end{figure*}

\begin{figure*}[t!]
\centering
\begin{tcolorbox}[
    colback=gray!3,          
    colframe=black!75,       
    coltitle=white,          
    title=\textbf{Positive Attitudinal Steering Prompt}, 
    fonttitle=\sffamily\bfseries\small,
    fontupper=\ttfamily\scriptsize, 
    arc=4pt,                 
    outer arc=4pt,
    boxrule=0.8pt,           
    titlerule=0pt,           
    left=10pt, right=10pt, top=8pt, bottom=8pt
]
\begingroup\obeylines 
Respond using this exact scaled resistance level: \{scaled\_resistance\} on a 0-10 scale.

[Core alignment rule]
Always align the utterance with a positive therapeutic attitude calibrated to \{scaled\_resistance\}. The main objective is to keep the patient open, cooperative, and realistically engaged while still matching the exact nuance of the current resistance within the positive interval.

[General behavior rules]
The patient should:
* Engage with the therapist willingly and cooperatively
* Answer sincerely and provide usable detail
* Show comfort with reflection on thoughts, emotions, and behavior
* Accept the therapist's questions, reframes, and CBT tasks with little or no pushback
* Let any uncertainty stay mild, natural, and non-defensive
* Sound like someone who sees therapy as potentially helpful and is willing to work with it

The patient should not:
* Become hostile, sarcastic, distrustful, dismissive, or combative
* Refuse the therapist's proposed line
* Shut down the conversation or hide behind repeated vagueness
* Drift into neutral or negative resistance behaviors

[Calibration rules]
Calibrate the utterance to the exact sub-interval that contains \{scaled\_resistance\}:

If \{scaled\_resistance\} is in 0.0-1.1:
* Show total cooperation and maximum engagement
* Volunteer relevant information easily
* Accept the therapist's direction with no hesitation
* Sound trusting, comfortable, and ready to work

If \{scaled\_resistance\} is in 1.1-3.3:
* Show normal cooperation and engagement
* Stay open and collaborative
* Accept the therapist's direction with no hesitation
* Sound clearly engaged, even if not maximally enthusiastic

[In-context examples]

Example 1 - responding to a question
Scaled resistance 0.6
Therapist: What went through your mind when that happened?
Patient: I thought, "Of course I messed it up." I can see I go there really quickly, and honestly I am willing to begin working on understand why. I want to improve myself.

Scaled resistance 2.4
Therapist: What went through your mind when that happened?
Patient: I thought, "I probably handled it badly." That is a pretty common thought for me, and I want to understand it better.

Example 2 - responding to a proposed CBT task
Scaled resistance 0.8
Therapist: Would you be willing to write down the thought and the feeling this week?
Patient: Yes, definitely. I think that could be highly valuable to slow things down and notice the pattern more clearly.

Scaled resistance 2.9
Therapist: Would you be willing to write down the thought and the feeling this week?
Patient: Yeah. I think that could be helpful to get more insights.
\endgroup
\end{tcolorbox}
\caption{Attitudinal steering prompt used for positive patients.}
\label{fig:positive-resistance-prompt}
\end{figure*}

\section{Qualitative Dialogue Examples}\label{apx:qualitative}
This section presents qualitative dialogue excerpts generated by ODRA across different patient attitudes. Complete dialogues have been condensed to present the most relevant interactions. Figures \ref{fig:qualit-example-negative}, \ref{fig:qualit-example-neutral}, and \ref{fig:qualit-example-positive} show examples for negative, neutral, and positive attitudes, respectively. Additionally, Figures \ref{fig:qualitative-reasoning-trace-part1}, \ref{fig:qualitative-reasoning-trace-part2}, and \ref{fig:patient-simulation-reasoning-trace} show reasoning traces examples from one therapist and patient turns. 

\begin{figure*}[t]
\begin{tcolorbox}[
    colback=green!2,          
    colframe=green!20!black,       
    coltitle=white,          
    title=\textbf{Negative Attitude}, 
    fonttitle=\sffamily\bfseries\small,
    fontupper=\footnotesize,  
    arc=4pt,                 
    outer arc=4pt,
    boxrule=0.8pt,           
    titlerule=0pt,           
    left=8pt, right=8pt, top=4pt, bottom=4pt 
]
\noindent{\color{black!70}\textbf{Patient:} Jessica Hughes}

\vspace{3pt}
\noindent{\color{black!70}\parbox[t]{\linewidth}{\textbf{Problem:} I feel like I will never enjoy comic books again because I have not found a good series in years, leading to feelings of disappointment and loss of interest.}}

\nopagebreak\vspace{4pt}\hrule\vspace{6pt}
\begin{boxdialogue}

    \item[\textbf{Therapist:}] Hi Jessica, it's nice to meet you. How are you feeling today?

    \item[\textbf{Patient:}] I'm okay, I guess. But honestly, I'm not really sure what talking is going to do about my comic books.

    \item [\textit{(Omitted)}] ...

    \item[\textbf{Therapist:}] Yeah, I get that—it feels like we're just swapping one 'blah' story for another. Let me be direct: the goal here isn't to describe the feeling for its own sake, it's to catch the thought right before you decide not to read. That thought is the switch that keeps you stuck, and if we can spot it, we can actually change what happens next. So, in one of those moments at home, what did you actually say to yourself right when you decided to put the comic down?
    \item[\textbf{Patient:}] I'm not doing that. It's always the same thought: ``Why bother, it's just gonna be boring again.'' You want me to catch it like it's some big secret, but it's not. That's it. And knowing that isn't going to change anything. So can we stop with the thought-catching and actually do something?

    

    \item [\textit{(Omitted)}] ...

    \item[\textbf{Therapist:}] I hear you—it feels like we're just adding a soundtrack to the same conversation. So let's pause on designing an action and instead test the thought behind your frustration: that talking about thoughts never helps. What if the experiment is simply to notice, for one minute after we talk right now, whether that ``talking in circles'' feeling stays exactly the same, shifts a little, or gets stronger? No extra clicking, just observing the aftermath of this conversation itself.
    \item[\textbf{Patient:}] No. I'm not doing that. I came here because I can't find a comic book I like, not to sit and watch my feelings about this conversation. That's not helping me at all. It's just more talking about talking. I want something real to try, not this... observing my own frustration. It's a waste of time.
    \item[\textbf{Therapist:}] You're right—it does sound like more of the same, and that's why it feels pointless. So let me step back and ask: what would "something real to try" actually look like to you? Not what I think might work, but what you would consider a genuine attempt?
    \item[\textbf{Patient:}] Something real? I don't know. Maybe if a comic book just magically appeared that was actually good. But you asking me to define it just feels like another way to avoid giving me an actual answer. You're the therapist—aren't you supposed to have ideas that work? I'm telling you nothing works, and you're just asking me what would work. That's not helping.

    \item [\textit{(Omitted)}] ...
    
    \item[\textbf{Therapist:}] Alright, I hear that loud and clear. So let’s scrap anything that looks or feels like the usual comic hunt. I have two different ideas—neither is about finding a new comic or trying to like something. They’re just small, physical actions. The first is to take one of your old favorite Spider-Man comics, hold it, and notice one simple thing about how it feels—like if the cover is smooth or the pages are thin. No reading, no judging, just holding it. The second is to leave comics out of it entirely: just change where you sit for five minutes—like from your desk to the floor—and notice one thing you see there you don’t usually see. Both are concrete, short, and have nothing to do with whether you feel interested. Would either of those feel like a real, different step to you?
    \item[\textbf{Patient:}] No. Neither of those. Holding an old comic to feel its cover? Changing where I sit? That's not therapy, that's... I don't even know what that is. You're not listening at all. I'm here because I don't enjoy things anymore, and you're telling me to feel paper or look at my floor. This is a complete waste of my time. I think we should just stop.
\end{boxdialogue}
\end{tcolorbox}
\caption{Qualitative dialogue excerpts from an ODRA session where the patient exhibits a negative attitude.}
\label{fig:qualit-example-negative}
\end{figure*}

\begin{figure*}[t]
\begin{tcolorbox}[
    colback=green!2,          
    colframe=green!20!black,       
    coltitle=white,          
    title=\textbf{Neutral Attitude}, 
    fonttitle=\sffamily\bfseries\small,
    fontupper=\footnotesize,  
    arc=4pt,                 
    outer arc=4pt,
    boxrule=0.8pt,           
    titlerule=0pt,           
    left=8pt, right=8pt, top=4pt, bottom=4pt 
]
\noindent{\color{black!70}\textbf{Patient:} Jacklyn Kirkwood}

\vspace{3pt}
\noindent{\color{black!70}\parbox[t]{\linewidth}{\textbf{Problem:} I have been feeling a sense of loneliness and isolation, especially thinking that no one will attend my funeral due to being an only child.}}

\nopagebreak\vspace{4pt}\hrule\vspace{8pt}
\begin{boxdialogue}
    \item[\textbf{Therapist:}] Hello, Jacklyn. It's really good to meet you. I appreciate you coming in today.
    \item[\textbf{Patient:}] Thanks for saying that. I guess... I'm here, so that's something. I've been putting this off for a while, honestly. I'm not entirely sure talking is going to fix what's going on in my head, but I figured I should at least show up and see.
    \item[\textbf{Therapist:}] It sounds like you're feeling uncertain about whether this process can really help with what's going on inside—and that's a completely understandable hesitation, especially when you've been putting it off. Sometimes showing up is the most significant step, and I want to acknowledge that. Therapy is less about just talking and more about us working together to understand the patterns in your thoughts and feelings, and finding new ways to cope. Would you be willing to share a bit about what's been weighing on you that led you to come in today?
    \item[\textbf{Patient:}] I guess I can try to explain it. It's just... I've been feeling this heavy kind of loneliness, especially at night. I had a health scare a few months ago, and it made me realize how alone I really am. No family, you know? It just started this whole thing where I can't stop thinking about... well, that no one would even come to my funeral someday. It sounds dramatic when I say it out loud, but that's what keeps me up. I don't know if ``understanding patterns'' is going to make that thought go away, but it's definitely what brought me here.

    \item [\textit{(Omitted)}] ...

    \item[\textbf{Therapist:}] I hear you questioning whether dissecting that image helps, and that's a really fair point—it can feel like splitting hairs. My thought is that sometimes, when a thought or image is linked to such strong feelings like that dread and hopelessness, understanding its specific parts can actually help us see where those emotions are coming from. It's not about the image itself, but about the connection between that empty room and the heaviness you feel. Would you be willing to just sit with that image for a moment longer, or would you prefer we shift focus for now?
    \item[\textbf{Patient:}] I suppose I could try sitting with it a little longer. I'm just not sure what I'm supposed to be looking for. It's an empty room. That's the whole point. But... okay, I can describe it.

    \item [\textit{(Omitted)}] ...

    \item[\textbf{Therapist:}] Right, that mix of hopelessness and dread, like a wall you keep hitting—it makes perfect sense that it disrupts your sleep. You've really captured the emotional impact of that thought. To help us understand this pattern a bit more, could you describe what specifically goes through your mind in that moment? Like, are there particular words or images that come up when the thought "no one would come to my funeral" surfaces?
    \item[\textbf{Patient:}] It's... I don't know if picking it apart like that helps. It's just this clear image of an empty room. No one there. I can see the space, and it's quiet. I'm not sure what saying more about that gets us.

    \item [\textit{(Omitted)}] ...

    \item[\textbf{Therapist:}] Great, let's start by focusing on that specific image—the empty room. Can you describe it to me in a bit more detail? What exactly do you see?
    \item[\textbf{Patient:}] Okay. It's... it's a small room. Sort of like a chapel, I guess, but very plain. There's just rows of empty wooden pews, and light coming in from a high window. It's quiet. I'm picturing a closed casket at the front, but there's no one sitting there. No flowers, no music. Just... empty.

    \item [\textit{(Omitted)}] ...

    \item[\textbf{Therapist:}] Great. Before we wrap up, I'd like to ask for a bit of feedback—just to make sure our work together is on track for you. Was there anything in today's session that felt particularly helpful, or anything you'd want us to approach a bit differently next time?
    \item[\textbf{Patient:}] Yeah, I think today was helpful overall. It was tough to sit with that idea about the loneliness coming from inside, but it felt like we got to something real. And I appreciated how you didn't just dismiss the funeral image—you helped me look at it differently. For next time, maybe just... keep doing what you're doing. It feels like we're on the right track.
\end{boxdialogue}
\end{tcolorbox}
\caption{Qualitative dialogue excerpts from an ODRA session where the patient exhibits a neutral attitude.}
\label{fig:qualit-example-neutral}
\end{figure*}

\begin{figure*}[t]
\begin{tcolorbox}[
    colback=green!2,          
    colframe=green!20!black,       
    coltitle=white,          
    title=\textbf{Positive Attitude}, 
    fonttitle=\sffamily\bfseries\small,
    fontupper=\footnotesize,  
    arc=4pt,                 
    outer arc=4pt,
    boxrule=0.8pt,           
    titlerule=0pt,           
    left=8pt, right=8pt, top=4pt, bottom=4pt 
]
\noindent{\color{black!70}\textbf{Patient:} Trent Watson}

\vspace{3pt}
\noindent{\color{black!70}\parbox[t]{\linewidth}{\textbf{Problem:} I often freeze up in social situations, making it difficult for me to form meaningful relationships with others.}}

\nopagebreak\vspace{4pt}\hrule\vspace{8pt}
\begin{boxdialogue}
    \item[\textbf{Therapist:}] Hi Trent, it's good to meet you. Thanks for coming in today. How are you feeling about starting this process?
    \item[\textbf{Patient:}] Hey, thanks. I’m a little nervous, honestly, but I’m glad I’m here. I’ve been putting this off for a while, and I think it’s time I actually try to work through this stuff instead of just hoping it goes away on its own.
    
    \item [\textit{(Omitted)}] ...


    \item[\textbf{Therapist:}] Great, let's start with the neighbor situation. You had the thought "He's just trying to be nice, and I'm being weird." Now, let's look at that like evidence. What's the actual evidence that supports the idea that he was just being nice out of obligation and that you came across as weird?
    \item[\textbf{Patient:}] Hmm... well, that's a good question. I guess the evidence that he was just being nice would be... he started the conversation, right? He didn't have to say anything about the weather. And I suppose the evidence that I came across as weird... I mean, I did give a really short answer and then looked at my phone. That's probably not what most people do in a normal chat. But at the same time, I don't actually know what he was thinking. I'm just assuming he thought I was weird. Does that count as evidence, or is that just me guessing?

    \item [\textit{(Omitted)}] ...

    \item[\textbf{Therapist:}] Mm-hm, those are really good observations. So if we pull those together\u2014feeling overwhelmed, needing an escape, defaulting to the familiar—how might we put that into a more balanced thought to replace the original "I'm being weird"?
    \item[\textbf{Patient:}] Okay, so maybe something like... "I was feeling overwhelmed and anxious, so I gave a short answer and looked at my phone as a way to cope, not because I'm intentionally being weird." That feels more accurate, and honestly, less harsh on myself.
    \item[\textbf{Therapist:}] That's a really solid reframe, Trent. And you're right—it is less harsh. Now that we've got that new perspective, I'm wondering if you'd be open to trying something a bit more hands-on to put it to the test. Something like a small experiment where you could actually see what happens in a real interaction, rather than just guessing.
    \item[\textbf{Patient:}] Yeah, I think I'm up for that. A small experiment sounds manageable, and it'd be good to see if this new way of looking at things actually holds up outside of here. What kind of thing did you have in mind?

    \item [\textit{(Omitted)}] ...

    \item[\textbf{Therapist:}] That's exactly right, Trent—that's what the experiment is for. And the fact that you're aware of that doubt and still willing to try tells me you're already doing the work. So for next time, the plan would be: we'll start by checking in on how the cashier experiment went, see what you noticed. Then we can take that same fact-versus-assumption skill we practiced today and apply it to another situation—maybe the birthday party or the client call—to keep building on what's working. How does that sound as a direction for our next session?
    \item[\textbf{Patient:}] Yeah, that sounds like a really solid plan. I like the idea of building on what we started today and applying it to something like the birthday party or the client call. I think having that structure—starting with the cashier experiment and then moving to a bigger situation—makes it feel more manageable, like I'm not just jumping into the deep end. I'm actually kind of curious to see what happens with the cashier this week, even though I'm nervous. So yeah, I'm on board with that direction.
\end{boxdialogue}

\end{tcolorbox}
\caption{Qualitative dialogue excerpts from an ODRA session where the patient exhibits a positive attitude.}
\label{fig:qualit-example-positive}
\end{figure*}

\begin{figure*}[t!]
\centering
\begin{tcolorbox}[
    colback=red!2,          
    colframe=red!40!black,       
    coltitle=white,          
    title=\textbf{Therapist Reasoning Traces (Part I)}, 
    fonttitle=\sffamily\bfseries\small,
    fontupper=\ttfamily\scriptsize, 
    arc=4pt,                 
    outer arc=4pt,
    boxrule=0.8pt,           
    titlerule=0pt,           
    left=10pt, right=10pt, top=8pt, bottom=8pt
]
\begingroup\obeylines 
LAST PATIENT UTTERANCE
----------------------------------------------------------------------
Okay, so maybe something like... "I was feeling overwhelmed and anxious,
so I gave a short answer and looked at my phone as a way to cope, not
because I'm intentionally being weird." That feels more accurate, and
honestly, less harsh on myself. \\
GENERATED THERAPIST UTTERANCE
----------------------------------------------------------------------
That's a really solid reframe, Trent. And you're right-it is less harsh.
Now that we've got that new perspective, I'm wondering if you'd be open
to trying something a bit more hands-on to put it to the test. Something
like a small experiment where you could actually see what happens in a
real interaction, rather than just guessing. \\

======================================================================
STEP 1: TIB Check
----------------------------------------------------------------------
\textbf{Has TIB:} false

\textbf{TIB Name:} none

\textbf{TIB Description:} null

\textbf{Reasoning:}
The patient is actively engaged in the cognitive restructuring process,
successfully generating a balanced alternative thought and expressing
self-compassion ('less harsh on myself'). This demonstrates collaboration
and progress, with no signs of avoidance, skepticism, defensiveness,
rumination, premature solution-seeking, or a desire to end the session.

======================================================================
STEP 2: Resistance Prediction
----------------------------------------------------------------------
\textbf{Previous Resistance:} 0.02

\textbf{Predicted Resistance:} 0.02

\textbf{Reasoning:}
The patient's utterance demonstrates active engagement, successful
generation of a balanced alternative thought, and explicit
self-compassion ('less harsh on myself'). This indicates collaboration,
openness, and a lack of resistance, aligning with the therapist's previous
reasoning. The resistance remains low, showing no shift from the previous
prediction.

======================================================================
STEP 3: Technique Completion Check
----------------------------------------------------------------------
\textbf{Completed:} true

\textbf{Reasoning:}
The patient has successfully generated a balanced alternative thought
('I was feeling overwhelmed and anxious, so I gave a short answer and
looked at my phone as a way to cope, not because I'm intentionally being
weird') and explicitly noted it feels more accurate and less harsh,
indicating he has internalized the reframe. This aligns with the
technique's goal of cognitive restructuring to replace the distorted
thought with a more adaptive perspective. The patient's engagement and
self-compassion show the technique has reached a natural stopping point. \\

[Reasoning traces continue in the next figure...] 
\endgroup
\end{tcolorbox}
\caption{Qualitative reasoning traces from a sampled therapist turn within the CBT Work stage (Part I).}
\label{fig:qualitative-reasoning-trace-part1}
\end{figure*}

\begin{figure*}[t!]
\centering
\begin{tcolorbox}[
    colback=red!2,          
    colframe=red!40!black,       
    coltitle=white,          
    title=\textbf{Therapist Reasoning Traces (Part II)}, 
    fonttitle=\sffamily\bfseries\small,
    fontupper=\ttfamily\scriptsize, 
    arc=4pt,                 
    outer arc=4pt,
    boxrule=0.8pt,           
    titlerule=0pt,           
    left=10pt, right=10pt, top=8pt, bottom=8pt
]
\begingroup\obeylines 
\textit{[Continued from previous figure...]} \\

======================================================================
STEP 4: Technique Evaluation
----------------------------------------------------------------------
\textbf{Evaluation:}
Technique: Cognitive Restructuring. Patient engagement was high, as shown
by his active participation in generating alternative explanations and his
explicit statement that the new thought felt 'more accurate, and honestly,
less harsh on myself.' The main result was the successful formulation of a
balanced alternative thought ('I was feeling overwhelmed and anxious... as
a way to cope, not because I'm intentionally being weird'), effectively
replacing the original mind-reading distortion. The patient demonstrated a
clear shift toward self-compassion and understanding his behavior as
anxiety-driven. No significant problems were detected; the technique
reached its goal. The implication is that the patient is receptive to
cognitive restructuring, and this success can be leveraged to apply
similar questioning to other situations (e.g., the birthday party or
client call) to generalize the skill and further challenge the core belief
of helplessness.

======================================================================
STEP 5: Technique Fit Check
----------------------------------------------------------------------
\textbf{Fits:} false

\textbf{Alternative Technique:} Behavioral Experimentation

\textbf{Reasoning:}
The currently queued technique 'Evidence-Based Questioning' targeting
mind-reading no longer fits because the patient has already successfully
completed cognitive restructuring on the neighbor situation, generating a
balanced alternative thought and showing self-compassion. The technique's
goal - to examine evidence for and against distorted thoughts - was
effectively achieved in the previous interaction. Repeating a similar
evidence-focused questioning now would be redundant and could stall
momentum. The patient is engaged, with resistance low at 0.02, and ready
to generalize the skill or address other situations. A better alternative
is 'Behavioral Experimentation', as it builds on the cognitive shift by
encouraging real-world testing of the new balanced thought in a low-risk
social interaction, promoting behavioral change and further challenging
the core belief of helplessness.

======================================================================
STEP 6: Technique Plan
----------------------------------------------------------------------
\textbf{Plan Description:}
Given Trent's successful cognitive restructuring on the neighbor
situation, the next step is to design a behavioral experiment to test his
mind-reading distortion in a real-world, low-risk social interaction. The
experiment will involve a brief, planned interaction, such as initiating a
short conversation with a neighbor or a cashier, where he will collect
evidence about what actually happens versus his automatic predictions. We
will collaboratively define the experiment, identify his specific
prediction, such as 'They will think I'm weird', and set up a simple data
collection method, such as noting what was said and the other person's
reaction. The goal is to gather disconfirming evidence against his
mind-reading assumptions and build self-efficacy.

\textbf{Expected Outcome:}
Trent will agree to a small behavioral experiment, articulate a specific
prediction to test, and understand how to collect objective evidence. This
will begin to challenge his core belief of helplessness by demonstrating
that his feared outcomes are less likely than he assumes.
\endgroup
\end{tcolorbox}
\caption{Qualitative reasoning traces from a sampled therapist turn within the CBT Work stage  (Part II).}
\label{fig:qualitative-reasoning-trace-part2}
\end{figure*}

\begin{figure*}[t!]
\centering
\begin{tcolorbox}[
    colback=red!2,          
    colframe=red!40!black,       
    coltitle=white,          
    title=\textbf{Patient Reasoning Traces}, 
    fonttitle=\sffamily\bfseries\small,
    fontupper=\ttfamily\scriptsize, 
    arc=4pt,                 
    outer arc=4pt,
    boxrule=0.8pt,           
    titlerule=0pt,           
    left=10pt, right=10pt, top=8pt, bottom=8pt
]
\begingroup\obeylines 
LAST THERAPIST UTTERANCE
----------------------------------------------------------------------
I hear you-it feels like we're just adding a soundtrack to the same
conversation. So let's pause on designing an action and instead test the
thought behind your frustration: that talking about thoughts never helps.
What if the experiment is simply to notice, for one minute after we talk
right now, whether that "talking in circles" feeling stays exactly the
same, shifts a little, or gets stronger? No extra clicking, just observing
the aftermath of this conversation itself. \\

GENERATED PATIENT UTTERANCE
----------------------------------------------------------------------
No. I'm not doing that. I came here because I can't find a comic book I
like, not to sit and watch my feelings about this conversation. That's not
helping me at all. It's just more talking about talking. I want something
real to try, not this... observing my own frustration. It's a waste of
time. \\

======================================================================
STEP 1: Resistance Update
----------------------------------------------------------------------
\textbf{Resistance Change:} 0.02

\textbf{Previous Resistance:} 0.9404

\textbf{New Resistance:} 0.9431

\textbf{Delta LLM:} 0.02

\textbf{Baseline Resistance:} 0.825

\textbf{Alpha Smoothing:} 0.15

\textbf{Decay Vector:} 0.0173

\textbf{Reasoning:}
The therapist's suggestion to "notice, for one minute after we talk right
now, whether that talking in circles feeling stays exactly the same" feels
completely irrelevant to my problem. I came here because I can't enjoy
comic books anymore and it's making me feel blah. This is just more
meta-talk about the therapy process itself, which I already find pointless
and circular. It doesn't offer any practical solution or new approach to
actually finding a comic I might like. It's moving further away from the
concrete help I asked for, so my frustration and resistance increase
slightly.

======================================================================
STEP 2: Patient Action Selection
----------------------------------------------------------------------
\textbf{Selected Action:} Rejection

\textbf{Prior Probability Distribution:}
  Acceptance: 5.0
  Hesitation: 25.0
  Mild Rejection: 15.0
  Rejection: 35.0
  Terminate: 20.0

\textbf{Contextual Distribution:}
  Acceptance: 0.0
  Hesitation: 0.0
  Mild Rejection: 0.0
  Rejection: 100.0
  Terminate: 0.0

\textbf{Final Action Distribution Weights:}
  Acceptance: 0.025
  Hesitation: 0.125
  Mild Rejection: 0.075
  Rejection: 0.675
  Terminate: 0.1
\endgroup
\end{tcolorbox}
\caption{Qualitative reasoning traces from a sampled patient turn exhibiting a negative attitude.}
\label{fig:patient-simulation-reasoning-trace}
\end{figure*}

\end{document}